\documentclass[preprint,10pt,3p]{elsarticle}

\usepackage{amssymb}
\usepackage{tikz}
\usepackage{mathtools}
\usepackage{natbib}
\usepackage{multirow}
\usepackage{color}
\usepackage{soul}
\usepackage{url}
\usepackage{hyperref}
\usepackage{eurosym}
\usepackage{graphicx}
\usepackage{caption}
\usepackage{xcolor}
\usepackage{amsmath}
\usepackage{subcaption}
\usepackage{framed}
\usepackage{multicol}
\usepackage{nomencl}
\usepackage{booktabs}
\usepackage{colortbl}
\usepackage{bm}
\usepackage[linesnumbered,ruled]{algorithm2e}
\usepackage{tikz}
\usetikzlibrary{shapes.geometric, arrows}
\tikzstyle{startstop} = [ellipse, minimum width=3cm, minimum height=1cm,text centered, draw=black]
\tikzstyle{io} = [trapezium, trapezium left angle=70, trapezium right angle=110, minimum width=3cm, minimum height=1cm, text centered, text width=8cm, draw=black]
\tikzstyle{process} = [rectangle, minimum width=3cm, minimum height=1cm, text centered, text width=5cm, draw=black]
\tikzstyle{decision} = [diamond, aspect=2, minimum width=2.5cm, minimum height=1cm, text centered, text width=3cm, draw=black]
\tikzstyle{arrow} = [thick,->,>=stealth]
\usepackage[edges]{forest}
\makenomenclature

\renewcommand{\nompreamble}{\begin{multicols}{2}}
\renewcommand{\nompostamble}{\end{multicols}}

\journal{Applied Energy}

\begin{document}
\begin{frontmatter}

\title{A Machine Learning-Based Framework for Clustering Residential Electricity Load Profiles to Enhance Demand Response Programs}

\author[label1]{Vasilis Michalakopoulos\corref{cor1}}
\ead{vmichalakopoulos@epu.ntua.gr}
\cortext[cor1]{Corresponding author}
\author[label1]{Elissaios Sarmas}
\author[label1]{Ioannis Papias}
\author[label1]{Panagiotis Skaloumpakas}
\author[label1]{Vangelis Marinakis}
\author[label1]{Haris Doukas}
\address[label1]{Decision Support Systems Laboratory, School of Electrical \& Computer Engineering, National Technical University of Athens, Greece}

\begin{abstract}
Load shapes derived from smart meter data are frequently employed to analyze daily energy consumption patterns, particularly in the context of applications like Demand Response (DR). Nevertheless, one of the most important challenges to this endeavor lies in identifying the most suitable consumer clusters with similar consumption behaviors. In this paper, we present a novel machine learning based framework in order to achieve optimal load profiling through a real case study, utilizing data from almost 5000 households in London. Four widely used clustering algorithms are applied specifically K-means, K-medoids, Hierarchical Agglomerative Clustering and Density-based Spatial Clustering. An empirical analysis as well as multiple evaluation metrics are leveraged to assess those algorithms. Following that, we redefine the problem as a probabilistic classification one, with the classifier emulating the behavior of a clustering algorithm,leveraging Explainable AI (xAI) to enhance the interpretability of our solution. According to the clustering algorithm analysis the optimal number of clusters for this case is seven. Despite that, our methodology shows that two of the clusters, almost 10\% of the dataset, exhibit significant internal dissimilarity and thus it splits them even further to create nine clusters in total. The scalability and versatility of our solution makes it an ideal choice for power utility companies aiming to segment their users for creating more targeted Demand Response programs.

\end{abstract}

\begin{keyword}
clustering methods \sep machine learning \sep electricity loads \sep residential consumers \sep demand response \sep smart-grid
\end{keyword}
\end{frontmatter}

\clearpage

\begin{table*}[!t]
\scriptsize
\begin{framed}
\nomenclature{CDI}{Clustering Dispersion Indicator}
\nomenclature{MDI}{Modified Dunn Index}
\nomenclature{SI}{Scatter Index}
\nomenclature{DBI}{Davies–Bouldin Index}
\nomenclature{MIA}{Mean Index Adequacy}
\nomenclature{SMI}{Similarity Matrix Indicator}
\nomenclature{AI}{Artificial Intelligence}
\nomenclature{VRC}{Variance Ratio Criterion}
\nomenclature{WCBCR}{Ratio of Within Cluster sum of squares to between Cluster Variation}
\nomenclature{CI}{Calinski-Harabasz Index}
\nomenclature{MSE}{Mean Square Error}
\nomenclature{SIL}{Silhouette Index}
\nomenclature{DI}{Dunn Index}
\nomenclature{SDI}{Standard Deviation Index}
\nomenclature{MDI}{Modified Dunn Index}
\nomenclature{VRC}{Variance Ratio Criterion}
\nomenclature{DB**}{Modified Davies-Bouldin Index}
\nomenclature{GS}{Gap Statistic}
\nomenclature{BIC}{Bayesian Information Criterion}
\nomenclature{DR}{Demand Response}
\nomenclature{SOM}{Self-Organizing Maps}
\nomenclature{FCM}{Fuzzy C-Means}
\nomenclature{FDL}{Modified Follow the Leader}
\nomenclature{PGM}{Probabilistic Generative Models}
\nomenclature{EM}{Expectation Maximization}
\nomenclature{IRC}{Iterative Refinement Clustering}
\nomenclature{MVM}{Minimum Variance Method}
\nomenclature{MFC}{Mixed Fuzzy Clustering}
\nomenclature{AP}{Affinity Propagation}
\nomenclature{MFC}{Mixed Fuzzy Clustering}
\nomenclature{xAI}{Explainable Artificial Intelligence}

\printnomenclature

\end{framed}
\end{table*}

\section{Introduction}
\label{sec:intro}

As renewable energy sources like solar and wind power become more and more prevalent, the intermittent nature of these sources necessitates a more flexible approach to energy consumption. A direct response to this changing energy landscape is the adoption of DR mechanisms. DR allows consumers, both residential and industrial, to actively participate in managing the electric grid by adjusting their electricity usage in response to specific signals, typically from utilities or grid operators \cite{4275494}. In times of peak demand or grid instability, consumers reduce or shift their electricity usage, lessening strain on the grid and preventing potential blackouts \cite{Bahrami2012}. DR not only promotes grid reliability but also contributes to environmental sustainability by synchronizing energy usage with the availability of renewable energy, reducing the need for backup power sources \cite{Parvania20131957}. Additionally, DR creates opportunities for cost savings, as participating consumers can benefit from incentives and lower electricity rates during non-peak hours \cite{Muratori20161108}. 

Power utilities and grid operators are increasingly using DR programs to help balance supply and demand, ultimately leading to lower electricity costs in wholesale markets and reduced retail rates \cite{en15051659}. Methods for involving consumers in DR programs include offering time-based pricing options such as time-of-use rates, critical peak pricing, variable peak pricing, real-time pricing, and critical peak rebates \cite{4275494}. For the effectiveness of the DR programs clusters of consumers are essential \cite{KAUR2022109236}. Clusters help streamline the deployment of DR programs by creating homogeneous groups of consumers. By grouping consumers with similar energy consumption patterns, preferences and resources, DR programs can tailor their approaches and incentives, maximizing participation and impact \cite{Qiu2023}. Clusters also encourage collaboration among prosumers facilitating peer-to-peer energy sharing, promoting local energy resilience, and fostering community engagement in sustainable energy practices \cite{Shi2023142, Qiu2021}.

Traditionally, power utilities defined consumer behaviors using rudimentary methods, relying on broad demographic categories and historical consumption data. While these methods provided some insights, they often proved to be insufficiently precise and adaptable as they didn't manage to capture the details and complexities that define energy consumption patterns. This lack of precision and agility hindered the ability of utilities to respond effectively to changing consumer preferences and emerging energy consumption trends and as a result their ability to provide tailored services and effective DR strategies was limited.

The advent of digitization and smart meters has revolutionized consumer behavior analysis. Real-time data from smart meters provide a granular understanding of energy consumption patterns, allowing for more accurate and timely insights \cite{6693793, Wang20193125}. For instance, analyzing time of day energy consumption, daily usage pattern stability, and actual volume of energy use can provide valuable insights into household energy consumption habits \cite{MOTLAGH201911, en11092235}. Moreover, artificial intelligence (AI), with its advanced analytics and machine learning capabilities, plays a pivotal role in transforming this data into actionable knowledge. AI algorithms can semi-automatically identify and refine clusters of consumers based on a multitude of factors, such as consumption habits, peak demand periods, and response preferences. Techniques like Hierarchical clustering, K-means, Fuzzy C-means and SOM uncover complex patterns in large datasets, facilitating the creation of more accurate and meaningful consumer clusters \cite{RAJABI2020109628}.
\par While various clustering techniques and methodologies have been proposed, a significant gap still remains in establishing a systematic approach to identify the ideal number of consumer clusters for the effective design of DR programs. Additionally, there is a need to gain a deeper understanding of the underlying factors contributing to the formation of these segments.

This paper seeks to address this critical gap in the existing literature by introducing a systematic, data-driven method for determining the optimal number of clusters when classifying electricity consumers for DR programs, using historical data as the basis. Through this contribution, our research aims to provide valuable practical insights to energy industry stakeholders and utility companies, ultimately facilitating the development of more efficient and tailored DR initiatives, for a greener and more sustainable future in the residential energy domain.

The contributions of our study are summarized as follows:

\begin{itemize}
    \item Implant thorough description of time-series data through comprehensive feature engineering, targeted for DR-programs. The pipeline is tested on 4438 houses for more than one year of data, thus showing the generality of our approach in terms of load and time.
    \item We utilize and compare four different clustering algorithms –namely K-means, K-medoids, hierarchical and based clustering– using various dissimilarity indexes. We assess them using DBI, CHI and silhouette score. However, to end up with the prevailing clustering algorithm, in terms of cluster meaning, we perform an additional similarity matrix and visual evaluation, considering the specific requirements of a practical DR policy, taking into consideration factors such as optimal number of clusters, similarity and dissimilarity.
    \item We propose a solution that is scalable over time, allowing the integration of a significantly large number of households into the existing classes within mere seconds. This remarkable speed is attributed to the exceptionally low inference response time of the classifier.
    \item Reinterpretation of the problem as a probabilistic classification one, leveraging xAI to gain deeper insights and build a trusted environment between the power utilities and Machine Learning solutions. This way, the power utilities can clarify the reason that drives the formation of each user group, giving reliability to the solution, bridging the gap between empirical knowledge and AI. 
\end{itemize}

The rest of the study is organized as follows. In section \ref{sec:litrev} a thorough literature review is presented. In section \ref{sec:method} the developed methodology is described. In section \ref{sec:casestudy} an experimental application based on a real case study is presented. Finally, concluding remarks are provided in section \ref{sec:conclusions}.

\section{Literature Review}
\label{sec:litrev}

\subsection{Related Work}

Numerous research papers in the literature have explored the comparative analysis of various clustering techniques within the domain of electricity consumer classification and load pattern segmentation. In Table \ref{tab:lit} we present some of the ML methods used for segmenting electricity load consumers.

\begin{table}
\tiny
\centering
 \caption{Literature review on ML methods used for clustering electricity load consumers.}
 \begin{tabular}{p{0.25\linewidth} p{0.05\linewidth} p{0.25\linewidth} p{0.25\linewidth} p{0.1\linewidth}} 
 \hline
 Title & Year & Models & Metrics & Citations \\ [0.5ex]
 \hline
Comparisons among clustering techniques for electricity customer classification & 2006 & FDL, Hierarchical, K-means, FCM, SOM & CDI, MDI, SI, DBI & \cite{1626400} \\ 
Overview and performance assessment of the clustering methods for electrical load pattern grouping & 2012 & K-means, FCM, Hierarchical, Follow-The-Leader & CDI, DBI, MDI, MIA, SI, SMI, VRC, WCBCR & \cite{CHICCO201268} \\
A clustering approach to domestic electricity load profile characterisation using smart metering data & 2015 & K-means, K-medoids, SOM & DBI & \cite{MCLOUGHLIN2015190} \\ 
Load pattern-based classification of electricity customers & 2004 & FDL, SOM & CDI, MIA & \cite{1295037} \\ 
Hopfield–K-Means clustering algorithm: A proposal for the segmentation of electricity customers & 2011 & Hopfield–Kmeans, Hopfield’s Recurrent Neural Network, K-means, SOM–K-Means, FDL, Hierarchical & MIA, CI, DBI & \cite{LOPEZ2011716} \\ 
Application of clustering algorithms and self organising maps to classify electricity customers & 2003 & Hierarchical, K-means, FCM, FDL & MIA, SMI, CDI, DBI & \cite{1304160} \\ 
Comparison of clustering approaches for domestic electricity load profile characterisation - Implications for demand side management & 2019 & K-means & SIL & \cite{YILMAZ2019665} \\ 
A comparative study of clustering techniques for electrical load pattern segmentation & 2019 & K-means, FCM, Hierarchical, PGM, SOM & MSE, SIL, DBI, MIA, WCBCR, DI & \cite{RAJABI2020109628} \\ 
Evaluating different clustering techniques for electricity customer classification & 2010 & K-means, Weighted Fuzzy Average K-means, FDL, SOM, Hierarchical & CDI, MIA & \cite{5484234} \\ 
Simulation Study on Clustering Approaches for Short-Term Electricity Forecasting & 2018 & Hierarchical & CI, C Index, Duda Index, Ptbiserial Index, DB Index, Frey Index, Hartigan Index, Ratkowsky Index, Ball Index, McClain Index, KL Index, SIL, DI, SDI & \cite{article} \\ 
A review on clustering of residential electricity customers and its applications & 2017 & K-means, K-medoids, K-medium, FCM, Adaptive K-means, Hierarchical, SOM, FDL, EM & CDI, DBI, MIA, SMI, DI, MDI, SIL, WCBCR, SI & \cite{8056062} \\ 
Efficient iterative refinement clustering for electricity customer classification & 2005 & IRC, Follow-The-Leader & CDI, MIA, SI, VRC & \cite{4524366} \\ 
A Novel Clustering Index to Find Optimal Clusters Size With Application to Segmentation of Energy Consumers & 2021 & K-means, Single Linkage & CI, DBI, DB**, GS, BIC, SIL, S\_Dbw & \cite{9072418} \\ 
Deriving the optimal number of clusters in the electricity consumer segmentation procedure & 2013 & MVM, FCM & WCBCR, BIC & \cite{6607329} \\ 
Analysing the segmentation of energy consumers using mixed fuzzy clustering & 2015 & MFC & CI, DBI, SIL & \cite{7338120} \\ 
Load Profile Based Electricity Consumer Clustering Using Affinity Propagation & 2019 & AP & SIL, CI, DBI, DI, WCBCR, CDI & \cite{8833693} \\ 
Application of clustering technique to electricity customer classification for load forecasting & 2015 & K-means, FCM & Proposed clustering validity indicator & \cite{7279510} \\

 \hline
 \end{tabular}
 \label{tab:lit}
\end{table}

Chicco et al. conducted a study \cite{1626400} that comprehensively compares different unsupervised clustering algorithms – hierarchical clustering, K-means, Fuzzy C-means (FCM), Modified Follow the Leader (FDL) and Self-Organizing Map (SOM) – in terms of their effectiveness in classifying electricity customers based on their consumption patterns. The findings of the study indicate that hierarchical clustering and FDL methods when run with the average distance linkage criterion offer a more detailed separation of clusters and electricity consumption behavior patterns compared to alternative clustering techniques. In the study by Amin Rajabi et al. \cite{RAJABI2020109628}, among the clustering methods considered, the K-means algorithm emerges as the most commonly utilized, followed by FCM, hierarchical clustering, and SOM.

The domain of smart meter data has also witnessed the application of clustering techniques. In the study \cite{6688044}, utility customer segmentation based on smart meter data is empirically examined. The researchers introduce statistical methods that leverage variability measurements to identify consumer segments of varying sizes, which can yield significant outcomes for tailored energy programs. They group the daily profiles of 85 customers using clustering techniques such as K-means, K-medoids, and SOM. This approach sheds light on the fact that predicting individual-level energy consumption is more feasible for customers with stable load profiles and lower variability, as opposed to those with erratic load patterns. Kwac et al. \cite{6693793} explore the segmentation of household energy consumption patterns using hourly data and propose the use of entropy to capture consumption variability. McLoughlin, et al. \cite{MCLOUGHLIN2015190} present another clustering approach for characterizing domestic electricity load profiles via smart metering data. Their study encompass a comparison of the effectiveness of three commonly used unsupervised clustering methods - K-means, K-medoids, and SOM - based on the Davies-Bouldin validity index. Subsequently, a multi-nominal logistic regression is employed to establish connections between profile classes and various factors such as dwelling type, occupants and appliance characteristics.

Additionally, Motlagh et al. address the clustering of residential electricity customers through load time series data \cite{MOTLAGH201911}. They propose a feature-based clustering algorithm that effectively tackles the challenges posed by the high dimensionality of such data. This approach involves transforming load time series into map models, which can be easily clustered, thereby mitigating the limitations associated with extreme dimensionality. By clustering customers based on their load patterns, valuable insights into customer preferences, usage patterns, and energy demand characteristics are gleaned. This facilitates the design of tailored electricity sales packages that align with the consumption patterns of each cluster, ultimately leading to enhanced customer satisfaction. Devijver et al. \cite{https://doi.org/10.1002/asmb.2453} explore the clustering of residential electricity consumers utilizing high-dimensional regression mixture models. This involves the calculation of clusters followed by the fitting of predictive models for each cluster.

Furthermore, several novel DL clustering algorithms customized for electricity customer classification and load pattern segmentation have been proposed. For example, \cite{LOPEZ2011716} introduces a clustering algorithm that combines the hopfield neural network and K-means algorithm to enhance electricity customer segmentation accuracy. The hopfield artificial neural network is utilized to eliminate the random generation of initial cluster centers. In another study, the Iterative Refinement Clustering (IRC) method merges desirable properties of hierarchical clustering and the Follow-the-Leader approach to achieve effective clustering outcomes \cite{4524366}. This method operates with a predetermined number of clusters and incorporates a specific mechanism to prevent the creation of empty clusters. Zarabie et al. \cite{8833693} assert the superior grouping capabilities of the Affinity Propagation Algorithm compared to other methods such as K-means, K-medoids, and spectral clustering in terms of grouping residential load profiles. These innovative approaches aim to improve the accuracy and efficiency of clustering techniques in the electricity domain.

At last, studies have explored the optimal number of clusters in the electricity consumer segmentation process. Nameer Al Khafaf et al. \cite{9072418} propose a clustering index derived from eigenvalue analysis on the correlation matrix of smart meter data to determine the optimal cluster size. To control the number of features, they also incorporate a genetic algorithm-based feature selection process. Similarly, \cite{6607329} suggests methods for optimizing cluster numbers by considering factors such as geographical characteristics, voltage level, and demographic features.

After conducting a review of the existing literature we identified significant gaps in the interpretation of formed clusters and the scalability of the methods employed. Many studies present segmentation methods that appear effective on a small scale but lack consideration for their applicability to larger populations. As power utilities increasingly seek to engage a broader range of consumers in DR initiatives, it is crucial to develop scalable segmentation methods that can support the complexity and diversity of the utility's consumer base. Additionally, while studies may successfully identify distinct consumer groups based on various characteristics such as consumption patterns and demographics, they often fall sort in explaining the reason why these clusters were formed and how they can translate into actionable strategies for utilities.

In this paper, we address the identified knowledge gaps in the literature by proposing an innovative solution that compares K-means, K-medoids, Hierarchical and Density-based Spatial clustering algorithms using DBI, CHI and SIL indexes. In order to determine the most suitable algorithm based on cluster meaning we conduct a similarity matrix and visual assessment with attention to DR needs. To provide a scalable solution we utilize a classifier for rapid household integration into clusters and xAI for detection of the reasons behind each user group's formation.

\section{Methodology}
\label{sec:method}

\begin{figure*}[tb] 
\centering
 \makebox[\textwidth]{\includegraphics[width=.9\paperwidth]{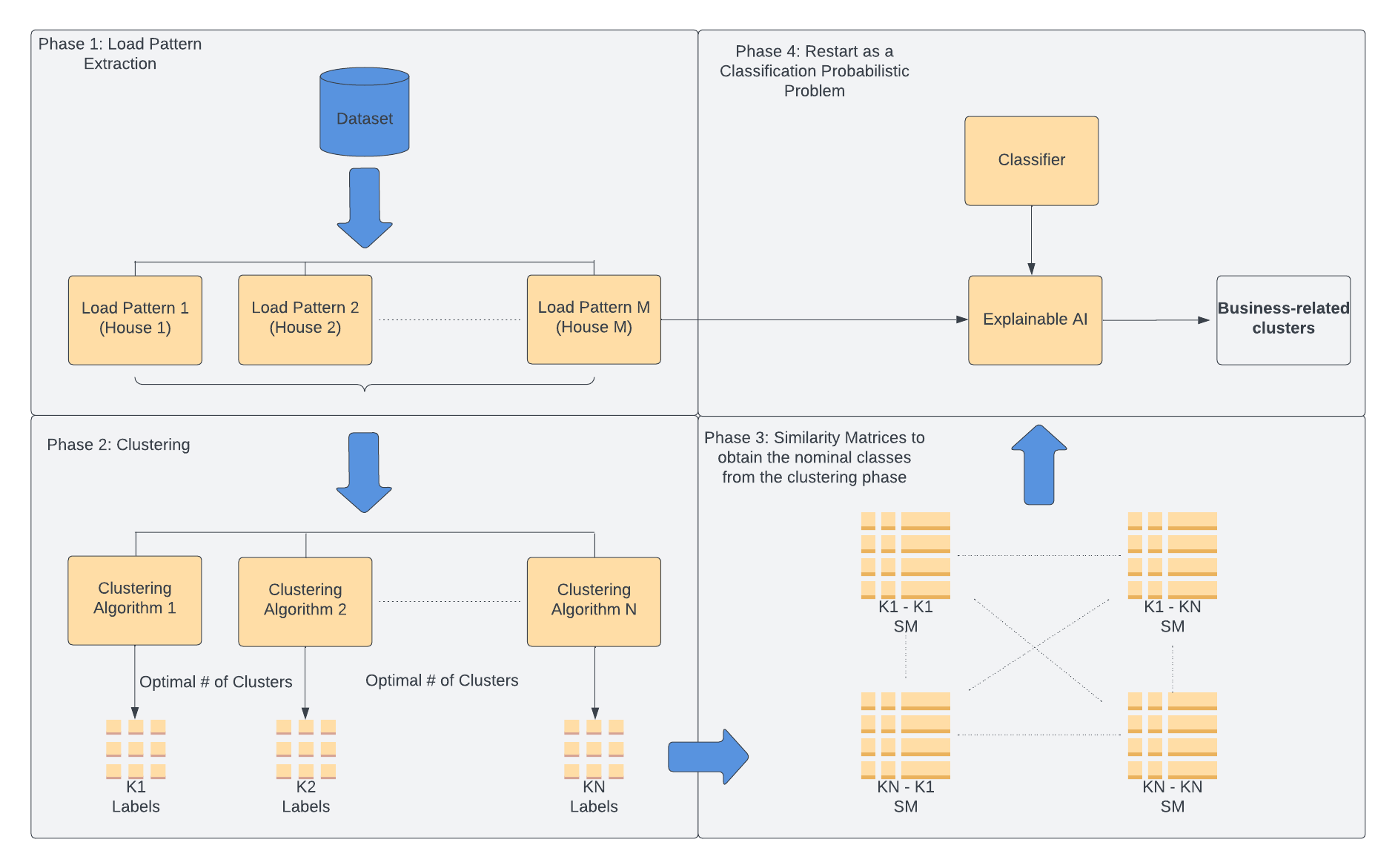}}
\caption{Proposed Methodology.}
\label{fig:Solution}
\end{figure*}

The proposed methodology is presented in Figure\ref{fig:Solution}. First, a detailed description of the first phase of the pipeline that extracts the features from the given data is issued. Following, clustering algorithms and the appropriate evaluation metrics are utilized to obtain the optimal number of clusters. In the next subsection, similarity matrices between the different algorithms are used to obtain the nominal clusters produced. On top of those clusters, a probabilistic classification function and xAI part are introduced as the final step of our proposed approach.

More specifically, the first phase of the process begins with load profiling of different residential households, capturing essential energy usage patterns and creating new features to enable data-driven decision-making. The obtained load profiles are then subjected to various clustering algorithms, utilizing evaluation metrics to determine the optimal number of clusters for accurate representation. To ensure the robustness of the results, similarity matrices are employed to assess the consensus among those clustering algorithms and produce the nominal clusters. Subsequently, the problem is reinterpreted as a probabilistic classification challenge and xAI techniques are used for the creation of additional specific clusters, failed to be identified in the previous step, facilitating deeper insights into energy consumption behavior. Through this seamless combination, stakeholders can gain comprehensive knowledge of household energy usage, identify energy saving opportunities, and implement effective DR programs to achieve environmental and economic benefits. Furthermore, the transparency facilitated by advanced xAI can bolster the confidence of power utilities and offer them a clear view of the key factors influencing the categorization of user groups, thus promoting well-informed decision-making processes.

\subsection{Load pattern extraction}

In this phase, energy consumption data from different residential buildings are initially collected and analyzed to create load profiles. Load profiling refers to the careful analysis of how energy is consumed or used over a specific period. It involves a detailed examination of patterns and behaviors in energy consumption during that defined timeframe. This examination involves identifying peak consumption periods, overall statistics of the consumption, and providing a detailed characterization of the overall usage patterns.

Through an extensive literature review, three distinct groups of attributes have been identified as the most important inputs for the subsequent clustering. The first group encompasses the raw time-series data, representing the average energy consumption for each examined house in 30-minute intervals throughout the day. The second group comprises features to pinpoint peak energy consumption during different periods throughout the day, namely the \texttt{peak early morning}, \texttt{peak morning}, \texttt{peak noon}, \texttt{peak evening}, \texttt{peak night}, and \texttt{peak late night}. More specifically for the mentioned time periods:

\begin{itemize}
  \item \texttt{peak early morning}: Represents the time frame from $05:00$ AM to $10:00$ AM, signifying the early morning peak in energy usage.
  \item \texttt{peak morning}: Encompasses the hours from $10:00$ AM to $02:00$ PM, identifying the morning peak in energy consumption.
  \item \textbf{peak noon}: Corresponds to the period between $02:00$ PM and $05:00$ PM, highlighting the midday peak in energy usage.
  \item \texttt{peak evening}: Covers the time span from $05:00$ PM to $09:00$ PM, indicating the evening peak in energy consumption.
  \item \texttt{peak night}: Extends from $09:00$ PM to $12:00$ AM, representing the nighttime peak in energy usage from $09:00$ PM to midnight.
  \item \texttt{peak late night}: Spans from $12:00$ AM to $05:00$ AM, indicating the late-night peak in energy consumption, covering the early morning hours.
\end{itemize}

The third group encompasses statistical descriptors, such as \texttt{mean} of the time-series, \texttt{std} (standard deviation), \texttt{min}, \texttt{max}, \texttt{25th percentile}, \texttt{50th percentile}, and \texttt{ 75th percentile}. These statistical measures offer a comprehensive view of data variability, and distribution characteristics, further enhancing our understanding of energy consumption patterns in the specific dataset. Leveraging these three distinct groups of attributes yields valuable insights into the intricacies of energy consumption behaviors, forming the cornerstone for subsequent clustering and classification algorithms.

\subsection{Clustering Algorithms}
\label{subsection:clustering}
Once the load profiling process has been completed, the next step involves applying various clustering algorithms, as an unsupervised learning technique, to group together similar energy consumption patterns within the data. Different clustering techniques, such as K-means, K-medoids, agglomerative and -based clustering, are employed to partition the data into distinct clusters. Each of these algorithms offers a unique approach to identifying patterns and structures within the energy consumption data, allowing for a comprehensive exploration of potential groupings. By applying multiple algorithms to the same data, it is possible to assess which one yields the best results in terms of cluster quality, inner-similarity and dissimilarity. The drawbacks and advantages of each algorithm, as applied in the proposed methodology, are described in the following paragraphs. 

\subsubsection{K-means}

The K-means algorithm \cite{kmeans} is one of the most widely used unsupervised learning technique in machine learning and it is commonly employed for clustering analysis, where a given dataset is divided into $k$ clusters. The minimazation of an objective function is the main goal of the K-means algorithm. This function represents the sum of the squared distances between each data point and its assigned cluster center.

Mathematically, the aforementioned function $J$ is defined in \eqref{eqn:second}:
\begin{equation}
\label{eqn:second}
    J = \sum_{i=1}^{k} \sum_{j=1}^{n} ||x_j - c_i||^2
\end{equation}

Where $J$ represents the objective function, $k$ is the number of clusters, $c_i$ is the centroid of cluster $i$, $x_j$ is a data point, and $||x_j - c_i||$ denotes the Euclidean distance between data point $x_j$ and cluster centroid $c_i$.

The K-means algorithm operates through an iterative process that involves the following steps:

\begin{enumerate}
    \item Initialization: Randomly select $k$ initial centroids or use a predefined initialization method.
    \item Assignment: Assign each data point to the nearest centroid, forming $k$ clusters.
    \item Update: Recalculate the centroids of each cluster as the mean of all data points in that cluster.
    \item Convergence: Repeat the assignment and update steps until convergence is achieved, i.e., when the centroids no longer change significantly or a predefined convergence criterion is met.
\end{enumerate}

The K-means algorithm is known for its simplicity, ease of implementation, and computational efficiency, which makes it suitable for analyzing large datasets. Over the years, various modifications and improvements have been proposed, including the use of different distance metrics, initialization methods, and convergence criteria, to enhance its performance in various applications.

\subsubsection{K-medoids}
The K-medoids algorithm \cite{kmedoids}, is a variation of the K-means algorithm. It is widely used for clustering analysis and has the advantage of being robust to outliers. K-medoids, unlike K-means, uses data points themselves (medoids) as cluster representatives.The medoid of a cluster is defined as the specific piece of data that exhibits the lowest average dissimilarity when compared to all other points within the same cluster.

The main objective of the K-medoids algorithm is to partition the dataset into $k$ clusters in a way that minimizes the total dissimilarity or cost. This cost is typically calculated as the sum of distances (using the Euclidean distance metric) between each data point and its assigned medoid.

Mathematically, the objective function $J$ for K-medoids can be represented in equation \eqref{eqn:first}:

\begin{equation}
\label{eqn:first}
    J = \sum_{i=1}^{k} \sum_{x_j \in C_i} d(x_j, m_i)
\end{equation}

Here, $J$ represents the objective function, $k$ is the number of clusters, $C_i$ is cluster $i$, $x_j$ is a data point in cluster $C_i$, $m_i$ is the medoid of cluster $C_i$, and $d(x_j, m_i)$ is the distance between data point $x_j$ and medoid $m_i$.

The K-medoids algorithm proceeds as follows:

\begin{enumerate}
    \item Initialization: Select $k$ initial medoids, either randomly or through a predefined method.
    \item Assignment: Assign each data point to the nearest medoid, forming $k$ clusters.
    \item Update: For each cluster, choose the data point as the new medoid that minimizes the total cost (dissimilarity) within that cluster.
    \item Convergence: Repeat the assignment and update steps until convergence is achieved. Convergence occurs when the assignment of data points to clusters and the selection of medoids no longer change significantly.
\end{enumerate}

The K-medoids algorithm is particularly useful when dealing with datasets that contain outliers, as it is less sensitive to outliers than K-means, due to its nature. It also offers flexibility in terms of the choice of distance metric, making it suitable for various types of data and applications. However, besides the advantages it has over K-means, it is more computationally expensive than the K-means algorithm, especially for large datasets. 

\subsubsection{Agglomerative hierarchical clustering}

Agglomerative clustering is a hierarchical clustering technique that initiates by assigning each data point to its own cluster. It is also known as the bottom-up approach or hierarchical agglomerative clustering (HAC). The assumption is that data points that are close to each other are more similar or related than data points that are farther apart. A dendrogram, a tree-like figure, is created by iteratively merging or splitting clusters based on a measure of similarity or distance between data points. Agglomerative clustering is quite flexible as it allows different linkage criteria to calculate the distance between pieces of data, and clusters by extension. Namely:

\begin{itemize}
    \item Single linkage: Measuring the distance between the nearest pair of points from different clusters.
    \item Complete linkage: Evaluating the distance between the farthest pair of points from different clusters.
    \item Average linkage: Calculating the average distance between all pairs of points from different clusters.
    \item Ward linkage : Minimizes the variance of the clusters being merged.
\end{itemize}

In mathematical terms, the agglomerative clustering algorithm can be described as shown in \eqref{eqn:third}:
\begin{equation}
\label{eqn:third}
    d_{ij} = \text{linkage}(C_i, C_j)
\end{equation}

where \(d_{ij}\) signifies the distance between clusters \(C_i\) and \(C_j\), and \textit{linkage} is the chosen linkage criterion.
Agglomerative clustering has several advantages, including its capability to handle non-linearly separable data and its interpretability through the dendrogram. Nevertheless, in the downsides section, it tends to be computationally intensive for sizable datasets and can exhibit sensitivity to noisy data.

\subsubsection{Density-based spatial clustering (DBSCAN)}

DBSCAN \cite{DBSCAN} is a density-based clustering algorithm that is particularly effective for identifying clusters with varying shapes and handling noisy data. It operates by defining clusters as dense regions of data separated by sparser areas, unlike the previous techniques.

The main objective of the DBSCAN algorithm is to partition the dataset into clusters such that each cluster is a dense region separated from other clusters by areas of lower point density. DBSCAN has several advantages, including its ability to discover clusters of varying shapes and sizes and its robustness to noise. It doesn't require specifying the number of clusters in advance, making it suitable for datasets with an unknown or variable number of clusters. Moreover, the choice of $\varepsilon$ (epsilon) and $MinPts$ parameters can significantly impact the clustering results. Tuning these parameters is often necessary to achieve the desired clustering outcome.

\subsection{Evaluation Metrics}
\label{subsection:evaluation}

To determine the optimal number of clusters for the data, the application of evaluation metrics becomes indispensable. Widely used metrics in the scientific literature, such as the silhouette score, DBI and the CHI were utilized for evaluating the quality of the clustering results. They assist in selecting the most suitable number of clusters that effectively capture and represent the inherent patterns in the data. By leveraging these evaluation metrics, researchers and practitioners can make informed decisions about the clustering outcomes and refine their analyses accordingly, enhancing the overall quality of the load profiling process.

\subsubsection{Davies-Bouldin index}

The Davies-Bouldin Index (DBI) \cite{DBI} is employed as a vital evaluation metric to determine the optimal number of clusters, denoted by \(K\). The index is defined as follows:
\begin{equation}
\label{eqn:fourth}
    DB = \frac{1}{K} \sum_{i=1}^{K} \max_{j \neq i} \left( \frac{S_i + S_j}{d_{ij}} \right) 
\end{equation}

where:
\begin{itemize}
    \item   \(S_i\) is the average distance between each data point within cluster \(i\) and its centroid.
    \item   \(d_{ij}\) is the distance between the centroids of clusters \(i\) and \(j\).
\end{itemize}

We perform the clustering algorithms with a range of cluster numbers, from 2 to 32, and calculate the DBI for each result. The optimal number of clusters is selected based on the clustering configuration that yields the lowest DBI. A lower index indicates better-defined clusters with reduced overlap, making it an appropriate criterion for identifying the optimal number of clusters. 

The DBI's sensitivity to cluster shape and distribution provides a comprehensive assessment of clustering quality, considering both compactness and separation. By employing this index as an evaluation metric, we aim to obtain an optimal clustering solution that best captures the underlying data structure and facilitates meaningful analysis of energy consumption patterns.

\subsubsection{Calinski-Harabasz index}

The Calinski-Harabasz index \cite{CBI} which was proposed in 1974 by Calinski T and Harabasz J, (also known as the Variance Ratio Criterion) is used as a significant evaluation metric to determine the optimal number of clusters, denoted by \(K\), in the clustering process. It is defined in \eqref{eqn:fifth}

\begin{equation}
\label{eqn:fifth}
    CH = \frac{B(K)}{W(K)} \times \frac{N - K}{K-1}
\end{equation}

where:

\begin{itemize}
    \item \(B(K)\) is the between-cluster sum of squares, representing the dispersion between the cluster centroids.
    \item \(W(K)\) is the within-cluster sum of squares, measuring the compactness of data points within clusters.
    \item \(N\) is the total number of data points.
\end{itemize}

Similar to the CHI, we apply the clustering algorithm with various cluster numbers, ranging from 2 to 32, and calculate the CHI for each result. The optimal number of clusters is selected based on the clustering configuration that yields the highest CHI. A higher index value signifies well-separated and compact clusters, making it an appropriate criterion for identifying the optimal number of clusters.

The CHI's ability to consider both inter-cluster and intra-cluster dispersion provides a comprehensive evaluation of clustering quality. By utilizing the CHI as an evaluation metric, we aim to obtain an optimal clustering solution that captures inherent patterns in the data and facilitates meaningful analysis of energy consumption patterns.

\subsection{Similarity matrices}

After obtaining the clustered energy consumption patterns, various algorithms are used for the clustering process. It is essential to evaluate the consistency and agreement between these different algorithms. To achieve this, similarity matrices are constructed to compare the labels assigned to data points in each cluster. A high similarity score indicates a strong consensus among the algorithms, ensuring potent results in the clustering outcomes and obtaining the nominal groups. On the other hand, having low similarity in cluster label indicates that this cluster might not be an accurate fit for the specific use case, despite the algorithms previously selecting it.

Adding to this, we employ t-SNE \cite{sne} (t-distributed Stochastic Neighbor Embedding) as a crucial tool to visualize and interpret our clustering results. t-SNE is a dimensionality reduction technique that excels in revealing the underlying structure of high-dimensional data by projecting it into a lower-dimensional space while preserving pairwise similarities. This reduction not only simplifies the representation but also enhances our ability to discern distinct clusters and patterns within the data. By utilizing t-SNE, we can transform complex multi-dimensional data into a two-dimensional space, like our case, where the clusters become visually apparent, enabling a more intuitive and insightful understanding of our clustering outcomes. This visualization not only assists in validating the quality of our clustering algorithm but also in generating valuable insights that enrich the broader context of our research, alongside the examination of similarity matrices.

\subsection{Classification \& xAI}

In this phase, additional clusters are created by incorporating business-related features and domain-specific information. After experimentation we selected the CatBoost Classifier, which was proposed by a team of engineers from Yandex in 2017 \cite{catboost} , and is employed to fit in the already given labels by the clustering algorithms, finding the probability of each value to be included in each cluster. CatBoost is an efficient classifier algorithm that uses gradient boosting on decision trees and handles categorical features in the data. Furthermore, categorical features are not handled during processing but are instead addressed during the training phase. Rather than employing binary substitution for categorical data, it utilizes random permutations and calculates a mean label value. To illustrate, similar class values are positioned ahead of the known ones in the permutation.

Additionally, while clustering algorithms are effective in grouping data points into clusters based on their similarities, they often struggle with generalization when applied to new, unseen data. This is where a classifier comes in handy. By training a classifier to perform the same function as the clustering algorithm, we enable it to make predictions on new data points and assign them to the appropriate clusters, thus extending the clustering model's ability to generalize. This combined approach not only enhances the accuracy and robustness of the clustering results but also provides a more versatile and adaptable framework for handling diverse data, making it a valuable tool in the case of electrical energy behavior analysis.

In traditional clustering methods, you might obtain clusters, but they might seem arbitrary or lack context. You have clusters, but you're not entirely sure why certain data points ended up together or how these clusters relate to your specific business goals. Due to this fact, xAI methods are employed to interpret the classifiers results and provide insights into the probabilities of each cluster. That way we can provide new clusters with actual business value and create them according to the objectives of the whole clustering approach.

\section{Case study}
\label{sec:casestudy}

\subsection{Data Set}
\label{sec:dataset}

\begin{figure*}[ht!] 
\centering
 \makebox[\textwidth]{\includegraphics[width=.75\paperwidth]{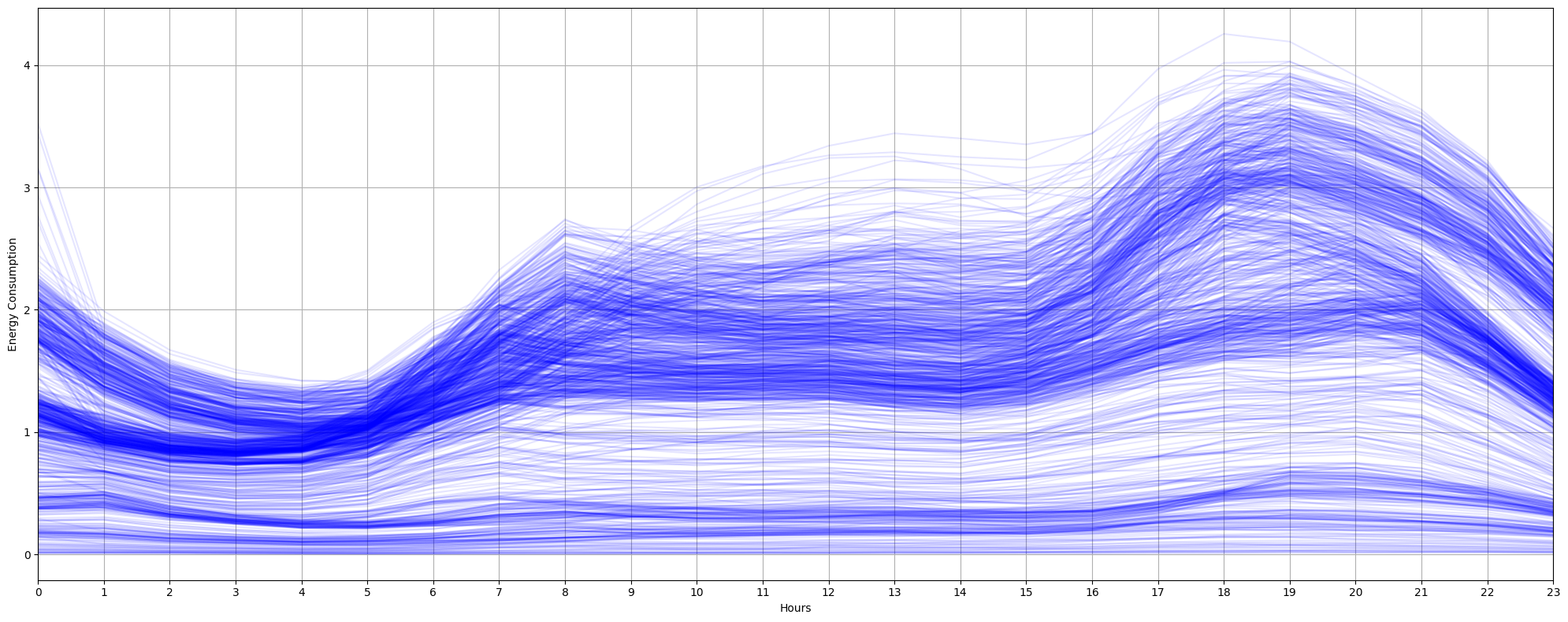}}
\caption{Average daily energy consumption for all househoulds }
\label{fig:dataset}
\end{figure*}

Open energy consumption data were used for the examined study that originated from 5,567 households in the city of London that span from November 2011 to February 2014 \footnote{\url{https://data.london.gov.uk/dataset/smartmeter-energy-use-data-in-london-households}}. The dataset captures energy readings at half-hour intervals and includes the energy consumption (kWh), the respective timestamp and a household identifier. Around 1100 of the households were exposed to dynamic Time-of-Use (dToU) energy prices throughout 2013. These customers received advance notifications of different price signals through Smart Meter IHD or mobile phone messages. This setup aimed to simulate potential future scenarios for effectively managing high renewable energy generation and leveraging high prices to ease the strain on local distribution grids during critical periods. In contrast, about 4500 customers who were not enrolled in dToU tariff schemes were considered more appropriate for examination due to their unfiltered data, which provides a better representation of real-world circumstances. Figure \ref{fig:dataset} presents the average energy consumption of the examined houses throughout the entire duration of data collection. As it can be seen, different groups of houses can be distinguished in terms of the volume of energy consumption. In addition, there are notable energy consumption peaks in the day, with most occurring early in the morning and in the evening, which can be attributed to the specific consumption pattern of the users for each household.

Real-world data often exhibit imperfections, such as incompleteness, inconsistency, inaccuracy, as well as the presence of errors, outliers, or missing values. Before delving into the data analysis, the data cleaning process was conducted to enhance data quality and uncover meaningful relationships within the dataset. Initially, data points containing missing or erroneous values were removed to ensure data integrity. Subsequently, a normalization procedure was employed, utilizing the MinMax method to scale all values within the standardized range of 0 to 1.0. This normalization step plays a pivotal role in mitigating the potential impact of varying data magnitudes on subsequent clustering analyses, thereby averting associated biases. The applied data preparation procedure aims to enhance the robustness of the subsequent clustering analyses by standardizing data scales, while also facilitating the application of distance-based metrics in data exploration.

\subsection{Experimental Design}
\label{sec:experimental}

After the cleaning process of the utilized data, the proposed pipeline is applied. The first step involves a careful analysis of the energy consumption patterns in the examined households. This analytical phase is pivotal in gaining insights into the temporal dynamics of energy usage for each consumer throughout the years.

Moving into the second part of the proposed framework, the four clustering algorithms, mentioned in \ref{subsection:clustering}, are applied to segment the data based on the features of each load profile. In the context of clustering analysis, determining the optimal number of clusters is a challenging task, as it typically cannot be precisely known in advance. Consequently, the various clustering algorithms are subjected to testing over a predefined range, spanning from 2 to 30 clusters. This broad range of cluster numbers is explored systematically to identify the most suitable or "best" number of clusters based on three evaluation metrics, including the Silhouette score, DBI and CHI, as mentioned in \ref{subsection:evaluation}.

\begin{figure*}[ht!]
    \centering
    \begin{subfigure}[b]{0.49\textwidth}
        \centering
        \includegraphics[width=\textwidth]{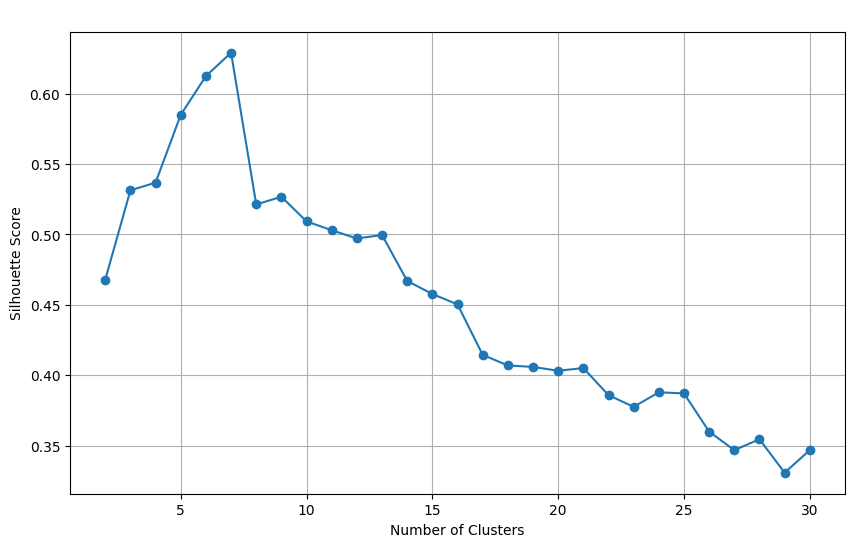}
        \caption{K-means}
        \label{fig:subfigure11}
    \end{subfigure}
    \begin{subfigure}[b]{0.49\textwidth}
        \centering
        \includegraphics[width=\textwidth]{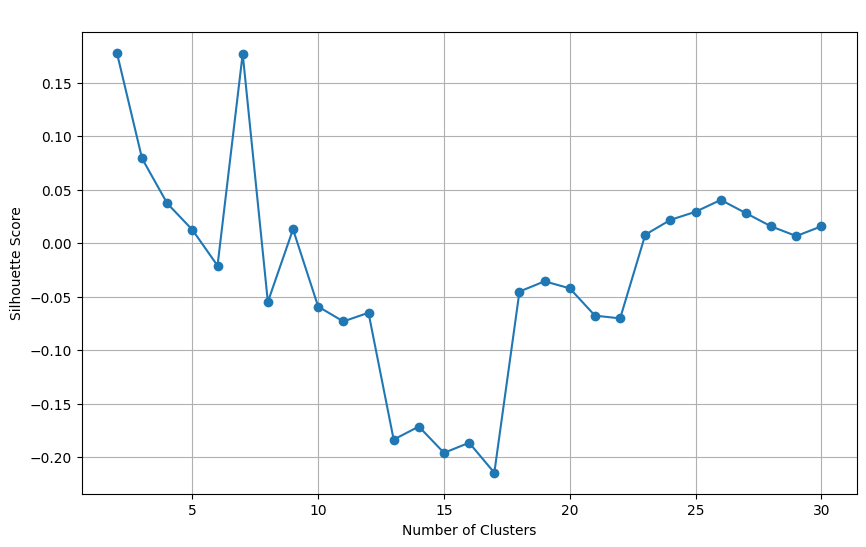}
        \caption{K-medoids}
        \label{fig:subfigure21}
    \end{subfigure}
    
    \vspace{0.5cm} 
    
    \begin{subfigure}[b]{0.49\textwidth}
        \centering
        \includegraphics[width=\textwidth]{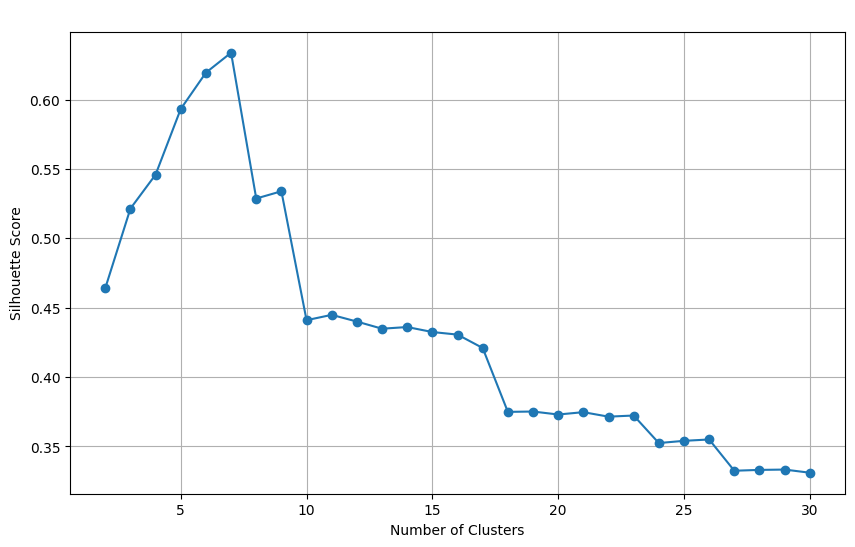}
        \caption{Agglomerative hierarchical}
        \label{fig:subfigure31}
    \end{subfigure}
    \caption{Silhouette score for every number of clusters and algorithm}
    \label{fig:shilouette_combined}
\end{figure*}

In Fig. \ref{fig:shilouette_combined}, we can see the difference in Silhouette scores for all the aforementioned algorithms, except DBSCAN, for every number of clusters tested. As can be seen from the diagram, the optimal number of clusters based on this metric is seven. The only exception can be found in \ref{fig:subfigure21}, where the number seven comes second in rank. Following this, the shilouette scores for the K-Medoids algorithm are too low compared to the other two algorithms and thus we do not consider this diagram quite representative to our case.

\begin{table}[ht!]
\footnotesize
\centering
\caption{Optimal number of clusters for each algorithm based on three evaluation indexes}
\begin{tabular}{|c|c|c|c|}
\hline
\textbf{Clustering Algorithm} & \textbf{Silhouette} & \textbf{DBI} & \textbf{CHI} \\
\hline
K-means & 7 & 7 & 7 \\
\hline
AgglomerativeClustering & 7 & 7 & 8 \\
\hline
K-medoids & 2 & 2 & 7 \\
\hline
\end{tabular}
\label{tab:experiment_results}
\end{table}

\label{sec:experiments}
\begin{figure*}[ht!] 
\centering
 \makebox[\textwidth]{\includegraphics[width=.7\paperwidth]{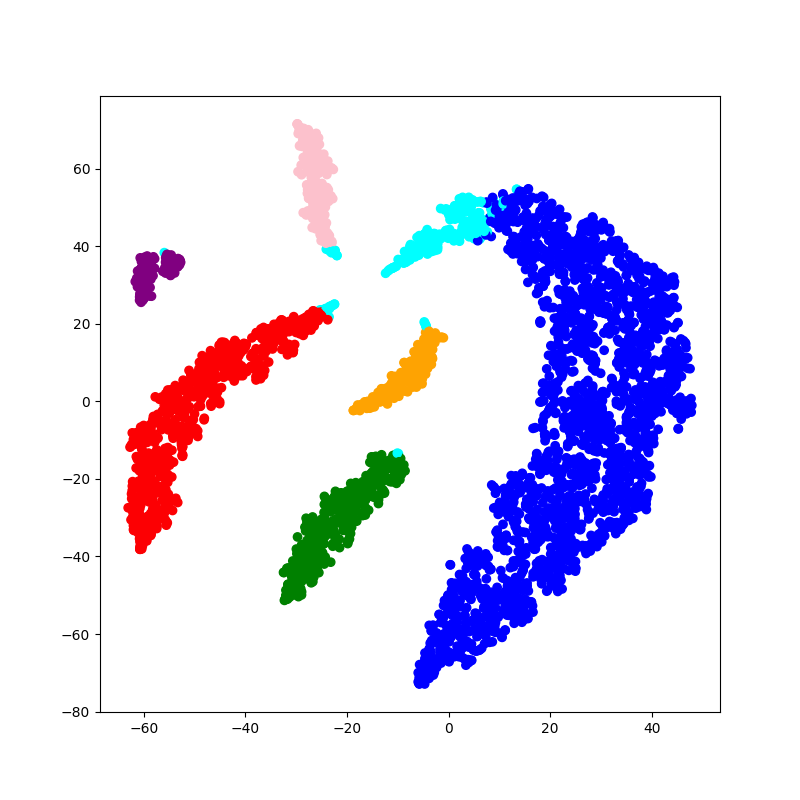}}
\caption{2D representation of the clusters for K-means labels}
\label{fig:tsne}
\end{figure*}

As previously mentioned, DBSCAN is a density-based clustering algorithm that does not necessitate users to predefine the number of clusters as an input parameter. Instead, DBSCAN dynamically determines the number of clusters based on the epsilon ($\varepsilon$) value and other criteria. Remarkably, during empirical testing with various epsilon values, it consistently identifies seven as the optimal number of clusters. Consequently, we select the cluster labels associated with the highest evaluation metrics across a range of epsilon values.

Adding to the previous, as we can see in table \ref{tab:experiment_results} most of the algorithms and their corresponding indexes agree that the number seven is the optimal one, in terms of load profile clustering, with the given features. Another useful visual that can be used to reduce the dimensions of the problem and give further insights into the clusters similarity and dissimilarity employing t-SNE, can be seen in Fig. \ref{fig:tsne}. As we can witness, the clusters are quite separated but improvements can be made. This is due to the fact, that some points that belong to different clusters are quite close in this diagram and can probably be a part of some other cluster or create a new, not already formulated one. Nevertheless, we anticipate confirmation from the similarity matrices to discern that the blue and red clusters are indeed the primary ones identified by the algorithms. Moreover, it is noticeable from this visualization that the cluster points highlighted in light blue exhibit the highest degree of dissimilarity. This observation suggests that there is room for enhancement in terms of improving cluster similarity within this specific cluster.

\begin{figure*}[ht!] 
\centering
 \makebox[\textwidth]{\includegraphics[width=.8\paperwidth]{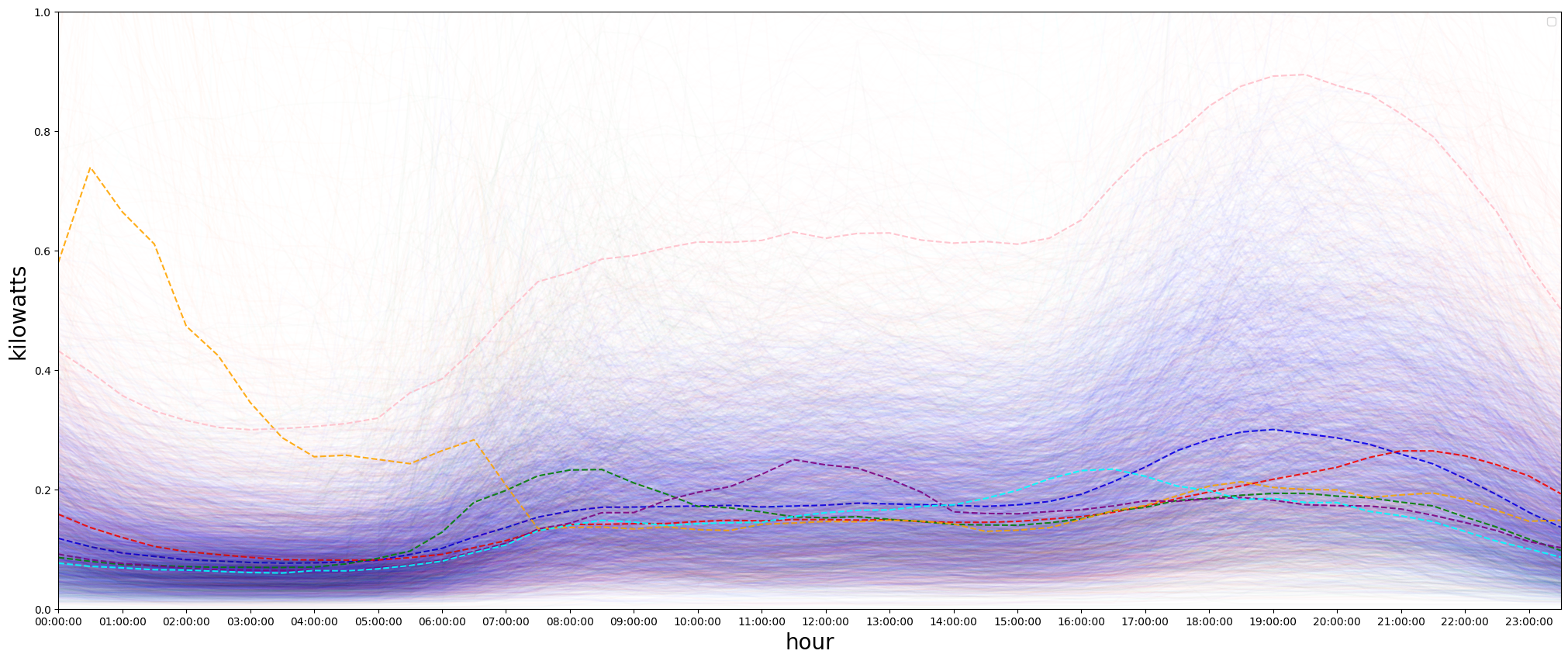}}
\caption{Median values for each cluster during the day}
\label{fig:clusters}
\end{figure*}

In Figure \ref{fig:clusters}, we present a visual representation of the clustering results obtained from our data analysis, on top of the timeseries data. Each line showcases the division of data points into seven distinct clusters, which were generated using the K-means algorithm. Each cluster is denoted by a unique color for visual clarity. It is evident that the data points within each cluster share common characteristics and exhibit a certain level of intra-cluster similarity, depending mostly on the daytime peaks of the timeseries. The distribution of data points across these clusters provides initial insights into the underlying structure of our dataset and the distinct cluster patterns it encompasses.

The decision to combine all clusters into a single figure serves as a powerful visualization strategy to showcase the antagonism (dissimilarity) between clusters. This choice is influenced by several key factors, with a primary emphasis on the incorporation of density information. Density serves as a fundamental metric, offering valuable insights into the concentration of data points within each cluster. The figure, by showcasing the density profiles of distinct clusters, aids in the identification of clusters exhibiting different levels of dispersion.

\begin{figure*}[ht!]
    \centering
    \begin{subfigure}[b]{0.45\textwidth}
        \centering
        \includegraphics[width=\textwidth]{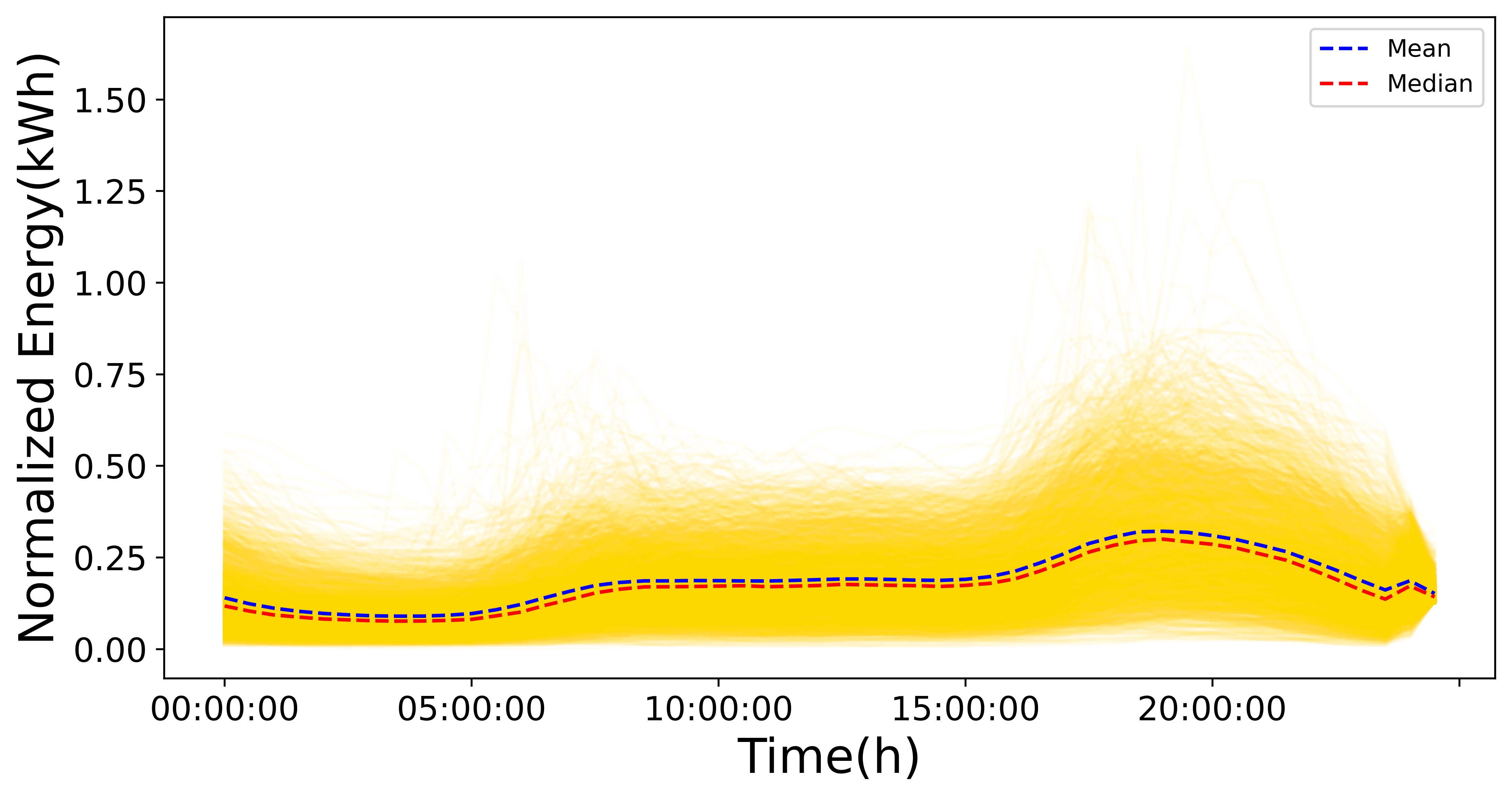}
        \caption{Cluster 0}
        \label{fig:subfigure12}
    \end{subfigure}
    \hfill
    \begin{subfigure}[b]{0.45\textwidth}
        \centering
        \includegraphics[width=\textwidth]{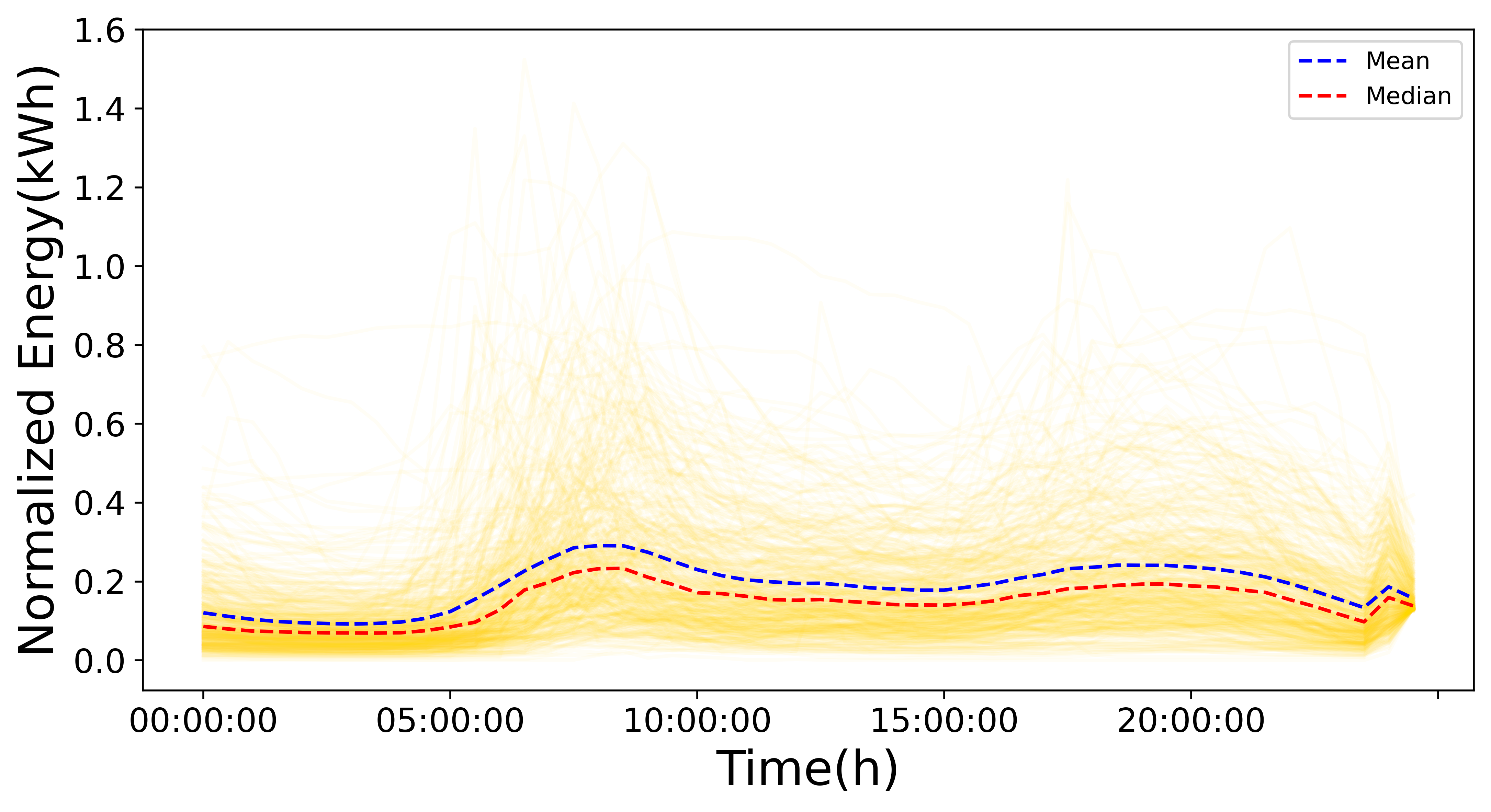}
        \caption{Cluster 1}
        \label{fig:subfigure22}
    \end{subfigure}
    \hfill
    \begin{subfigure}[b]{0.45\textwidth}
        \centering
        \includegraphics[width=\textwidth]{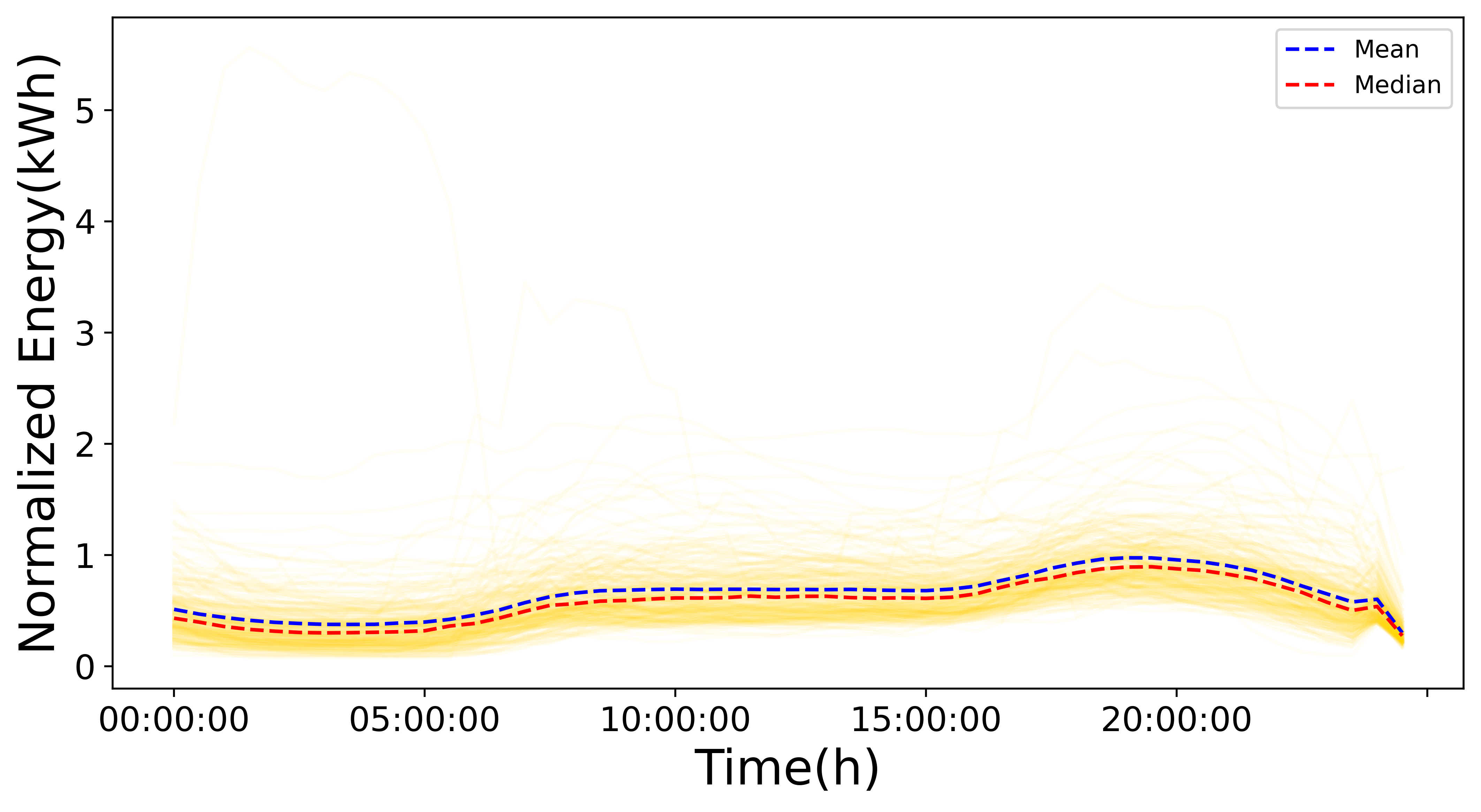}
        \caption{Cluster 2}
        \label{fig:subfigure32}
    \end{subfigure}
    \hfill
    \begin{subfigure}[b]{0.45\textwidth}
        \centering
        \includegraphics[width=\textwidth]{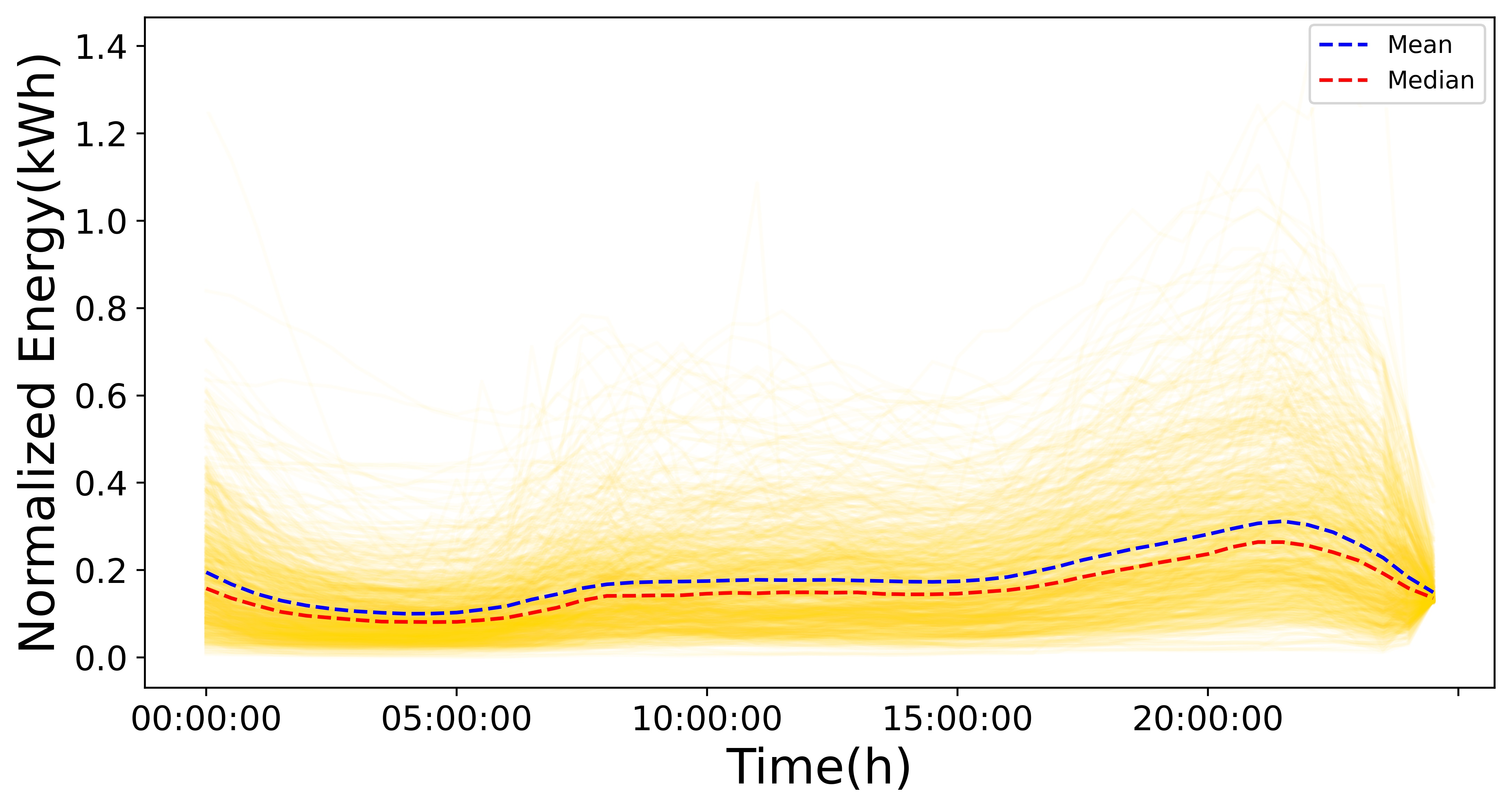}
        \caption{Cluster 3}
        \label{fig:subfigure4}
    \end{subfigure}
    
    \vspace{0.5cm} 
    
    \begin{subfigure}[b]{0.45\textwidth}
        \centering
        \includegraphics[width=\textwidth]{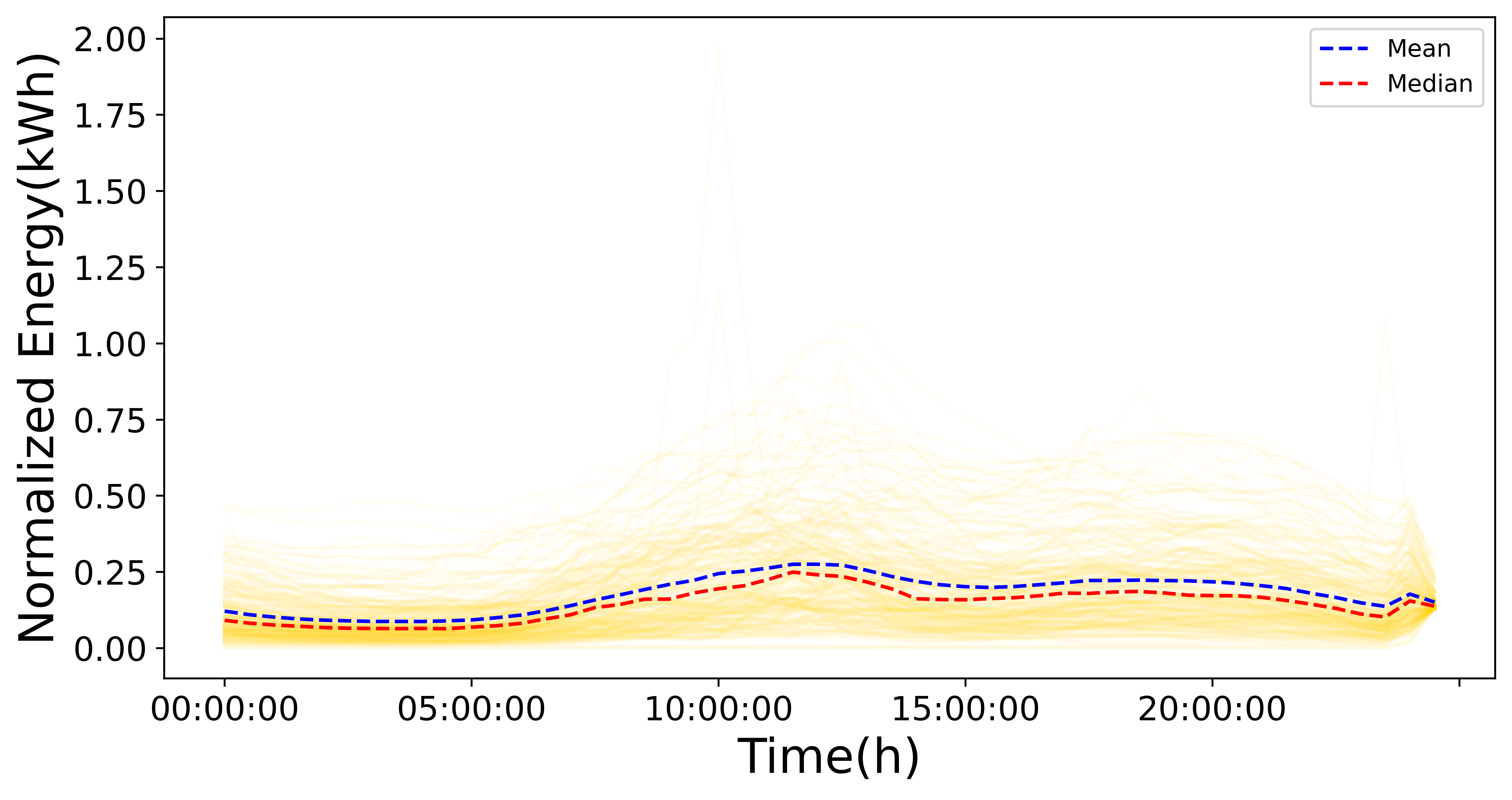}
        \caption{Cluster 4}
        \label{fig:subfigure5}
    \end{subfigure}
    \hfill
    \begin{subfigure}[b]{0.45\textwidth}
        \centering
        \includegraphics[width=\textwidth]{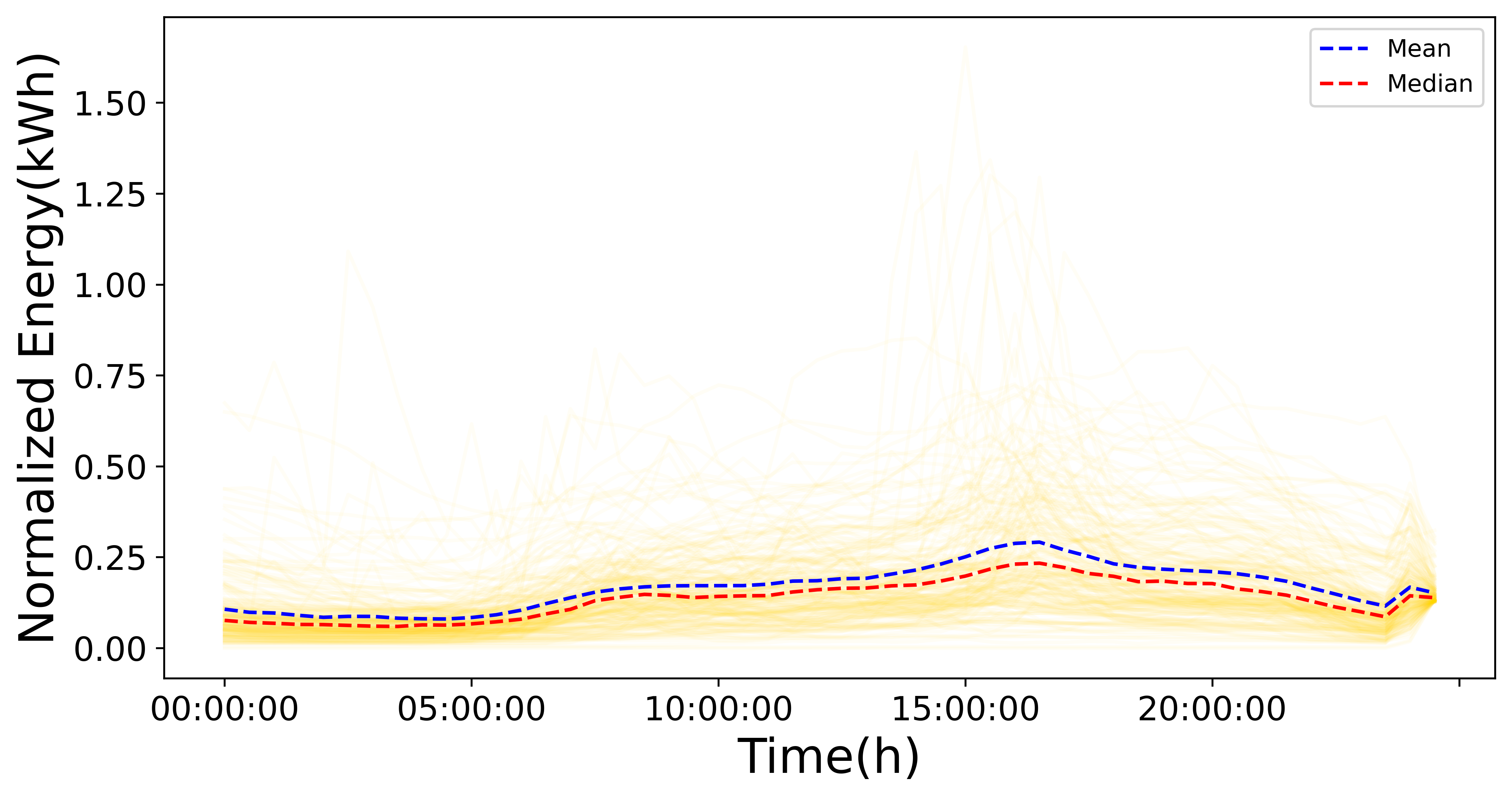}
        \caption{Cluster 5}
        \label{fig:subfigure6}
    \end{subfigure}
    \hfill
    \begin{subfigure}[b]{0.45\textwidth}
        \centering
        \includegraphics[width=\textwidth]{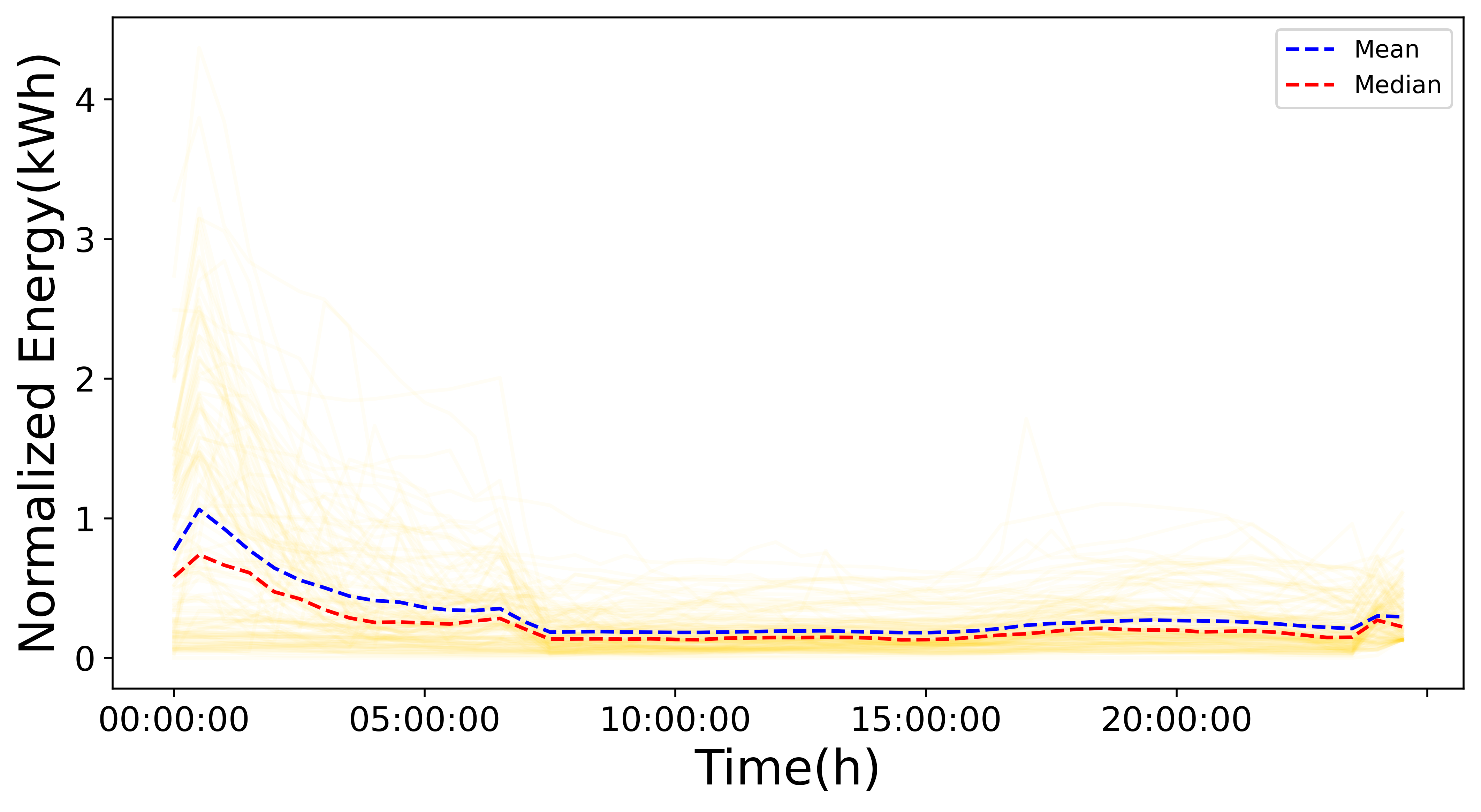}
        \caption{Cluster 6}
        \label{fig:subfigure7}
    \end{subfigure}
    
    \caption{Cluster Visualization}
    \label{fig:cluster_combined}
\end{figure*}

Adding to the previous, Figure \ref{fig:cluster_combined} presents a comprehensive visualization of all clusters, each displayed in separate subfigures, accompanied by essential statistical measures such as density, mean, and median. This integrated representation offers a holistic view of the clustering results, enabling a better understanding of each cluster. Moreover, the inclusion of mean and median values for each cluster's features further enhances the interpretability of the clusters. The mean provides a measure of central tendency, indicating the average behavior of data points within a cluster, while the median offers insights into the distribution's central point, the starting point for every center-based clustering algorithm.

Moving on to the similarity matrices between the algorithms, Table \ref{tab:comparison-similarity-matrices} illustrates the comparative performance of each clustering algorithm. Each row and column represent a cluster, and the values within the matrix reveal the degree of similarity between these clusters for the corresponding algorithms in comparison. Notably, as can be seen in Table \ref{tab:similarity-matrix-kmeans-agglomerative} , clusters 0 and 1 in the K-means vs. Agglomerative Clustering comparison exhibit a high degree of similarity, suggesting consistent and robust performance across these algorithms for these specific clusters. However, Cluster 4 in the same comparison shows a significant dissimilarity, indicating challenges in achieving agreement.

Similarly, the comparison between K-means and K-medoids reveals strong similarities in clusters 1 and 4, but also highlights variations in other clusters, as can be seen in Table \ref{tab:similarity-matrix-kmeans-kmedoids}. This divergence underscores the inherent differences in these two methods, with K-means prioritizing cluster centroids and K-medoids focusing on medians, leading to variations in cluster assignments. As expected from the low validation indexes scores produced by K-medoids, it is obvious that the latter is not suitable for our case and it will no longer be under consideration.

Table \ref{tab:similarity-matrix-agglomerative-dbscan} depicts the assessment of Agglomerative Clustering against DBSCAN, where distinct patterns of cluster formation emerge. The two algorithms agree almost in every data point for clusters 0,1 and 2, showing again that those are the nominal classes as previously shown in the other tables. To the contrary, classes 3 of Agglomerative and 6 of DBSCAN exhibit significant disagreement on data points between the algorithms. his discrepancy suggests that there is indeed room for improvement.

\begin{table}[ht]
\small
\centering
\caption{Comparison of Similarity Matrices}
\label{tab:comparison-similarity-matrices}

\begin{minipage}{0.35\textwidth}
\centering
\begin{subtable}{\linewidth}
\centering
\caption{K-means vs. Agglomerative Clustering}
\label{tab:similarity-matrix-kmeans-agglomerative}
\begin{tabular}{ccccccccc}
\toprule
 & 0 & 1 & 2 & 3 & 4 & 5 & 6 \\
\midrule
0 & 2442 & 0 & 0 & 0 & 0 & 0 & 0 \\
1 & 0 & 0 & 485 & 0 & 0 & 0 & 0 \\
2 & 0 & 0 & 0 & 0 & 417 & 0 & 0\\
3 & 0 & 0 & 0 & 0 & 0 & 239 & 0\\
4 & 111 & 9 & 6 & 123 & 0 & 0 & 0\\
5 & 0 & 399 & 0 & 0 & 0 & 0 & 0\\
6 & 0 & 0 & 0 & 1 & 0 & 0 & 206\\
\bottomrule
\end{tabular}
\end{subtable}

\vspace{0.5cm} 

\begin{subtable}{\linewidth}
\centering
\caption{K-means vs. K-medoids}
\label{tab:similarity-matrix-kmeans-kmedoids}
\begin{tabular}{cccccccc}
\toprule
 & 0 & 1 & 2 & 3 & 4 & 5 & 6 \\
\midrule
0 & 557 & 223 & 509 & 660 & 0 & 242 & 251\\
1 & 0 & 0 & 0 & 0 & 475 & 10 & 0\\
2 & 0 & 0 & 0 & 0 & 414 & 3 & 0\\
3 & 0 & 0 & 0 & 0 & 235 & 4 & 0\\
4 & 0 & 0 & 0 & 0 & 4 & 245 & 0\\
5 & 0 & 0 & 0 & 0 & 396 & 3 & 0\\
6 & 0 & 0 & 0 & 0 & 205 & 2 & 0\\
\bottomrule
\end{tabular}
\end{subtable}
\end{minipage}%
\hfill
\begin{minipage}{0.35\textwidth}
\centering
\begin{subtable}{\linewidth}
\centering
\caption{Agglomerative Clustering vs. DBSCAN}
\label{tab:similarity-matrix-agglomerative-dbscan}
\begin{tabular}{ccccccccc}
\toprule
 & 0 & 1 & 2 & 3 & 4 & 5 & 6 \\
\midrule
0 & 2553 & 0 & 0 & 0 & 0 & 0 & 0 \\
1 & 0 & 126 & 0 & 0 & 278 & 0 & 4 \\
2 & 0 & 0 & 0 & 0 & 0 & 0 & 491 \\
3 & 81 & 0 & 0 & 12 & 3 & 2 & 26 \\
4 & 0 & 0 & 416 & 0 & 0 & 0 & 1 \\
5 & 0 & 0 & 0 & 238 & 0 & 0 & 1 \\
6 & 0 & 0 & 0 & 0 & 0 & 206 & 0 \\
\bottomrule
\end{tabular}
\end{subtable}

\vspace{0.5cm} 

\begin{subtable}{\linewidth}
\centering
\caption{K-means vs. DBSCAN}
\label{tab:similarity-matrix-kmeans-dbscan}
\begin{tabular}{cccccccc}
\toprule
 & 0 & 1 & 2 & 3 & 4 & 5 & 6\\
\midrule
0 & 2442 & 0 & 0 & 0 & 0 & 0 & 0 \\
1 & 0 & 0 & 0 & 0 & 0 & 0 & 485\\
2 & 0 & 0 & 416 & 0 & 0 & 0 & 1\\
3 & 0 & 0 & 0 & 238 & 0 & 0 & 1\\
4 & 192 & 4 & 0 & 12 & 7 & 1 & 33\\
5 & 0 & 122 & 0 & 0 & 274 & 0 & 3 \\
6 & 0 & 0 & 0 & 0 & 0 & 207 & 0\\
\bottomrule
\end{tabular}
\end{subtable}
\end{minipage}

\end{table}

In our comparative analysis of Agglomerative Clustering, K-means, and DBSCAN ,using the provided similarity matrices, intriguing insights into cluster performance have emerged. Above all, while Agglomerative Clustering and DBSCAN exhibit alignment in certain clusters, such as cluster 2, 1 and 0, where they agree strongly, K-means often deviates from this consensus, particularly in cluster 4. This divergence highlights opportunities for improvement in the cluster results. For instance, clusters where K-means does not concur with other algorithms may benefit from further refinement. Conversely, the regions where all three algorithms agree, such as cluster 0,1,2,3 and 6 of K-means, provide a stable foundation for clustering analysis. On the other hand, clusters 4 and 5 indicate signs that further improvements are needed.

\subsection{Results}
\label{sec:results}

In the previous chapter, after comparing the clustering algorithms, data points have been individually labeled based on the outcomes obtained from each algorithm. Now that we have reached the final phase of our proposed methodology, we introduce the integration of a CatBoost classifier as a pivotal component of our analytical pipeline. This classifier is strategically employed to complement the existing clusters generated by our clustering algorithms. The classifier always learns from the clustering algorithm perfectly and thus it behaves exactly like the corresponding unsupervised learning technique. 

The labeled data sets are split into train and test parts, with an 80-20 split ratio. Then, test parts are hidden from the models during the training phase. In order to abtain class probabilities and enable probabilistic class interpretation, the Catboost's output is subject to a probability calibration process. In addition to making class predictions, this calibration procedure calculates the likelihood associated with each respective label through the classification process. These probabilities, computed for each data point, serve to provide a measure of confidence in the prediction outcomes. This application is made to add a new function to the clustering algorithms to return the probability value for each data point that is decided to belong to a certain cluster.

Now for the baseline of our approach we use the labels produced by the K-means algorithm as the training and test set, in order to see the accuracy of the classifier. From Table \ref{tab:catboost-metrics}, we can see that Catboost fits almost perfectly to the given data for every cluster.
In this case, while it is evident that Catboost's near-perfect fit to the data may be indicative of over fitting, it aligns well with our objectives, as we need the classifier to be mimicking the clustering algorithm.The only difference is that the user groups, will be mentioned as classes instead of clusters from now on, due to the change in problem formulation.

\begin{table}[ht]
    \centering
    \caption{CatBoost Algorithm Performance Metrics per class}
    \begin{tabular}{ccccc}
        \toprule
        Classes & Precision & Recall & F1 Score & \\
        \midrule
        Class 0 & 0.99 & 1.00 & 1.00 \\
        Class 1 & 1.00 & 0.99 & 0.99 \\
        Class 2 & 1.00 & 1.00 & 1.00 \\
        Class 3 & 1.00 & 1.00 & 1.00 \\
        Class 4 & 0.95 & 0.91 & 0.93 \\
        Class 5 & 1.00 & 1.00 & 1.00 \\
        Class 6 & 1.00 & 1.00 & 1.00 \\
        \bottomrule
    \end{tabular}
    \label{tab:catboost-metrics}
\end{table}

So from now on, we have a classifier that behaves almost identically to a clustering algorithm. Utilizing widely recognized metrics like Precision, Recall, and F1 Score, we can discern that the CatBoost algorithm exhibits remarkable performance across various classes. However, it's noteworthy that in most classes, including Class 0, Class 1, Class 2, Class 3, Class 5, and Class 6, the algorithm achieves exceptionally high Precision, Recall, and F1 Score values, suggesting that it is adept at identifying positive instances with a high degree of accuracy and comprehensively capturing relevant instances. This performance pattern means that it over fits to the existing data, as the algorithm seems to act nearly identically across most classes. This is exactly what we were aiming for, as mentioned earlier.Notably, in Class 4, we observe a deviation from this trend, where all the metric values are slightly lower, hinting at a divergence in algorithm behavior for this particular Class.

\begin{figure*}[ht!] 
\centering
 \makebox[\textwidth]{\includegraphics[width=.5\paperwidth]{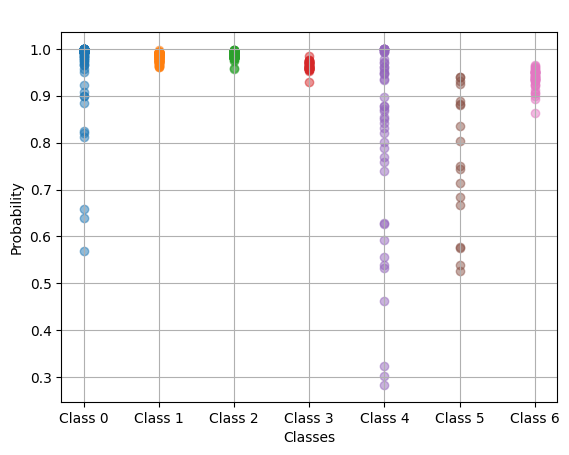}}
\caption{Probabilities for the predicted classes}
\label{fig:proba_all}
\end{figure*}

As can be seen in Figure \ref{fig:proba_all}, the only classes that hold room for improvement, except three points in class 0, are the ones we mentioned in our similarity matrices explanation, classes 4 and 5. It is obvious that instances assigned to classes with predictive probabilities falling below the threshold of 0.8 not only appear unsuitable for their designated class but also demonstrate limited suitability for the alternative existing classes.

\begin{figure*}[ht!] 
\centering
 \makebox[\textwidth]{\includegraphics[width=0.6\paperwidth]{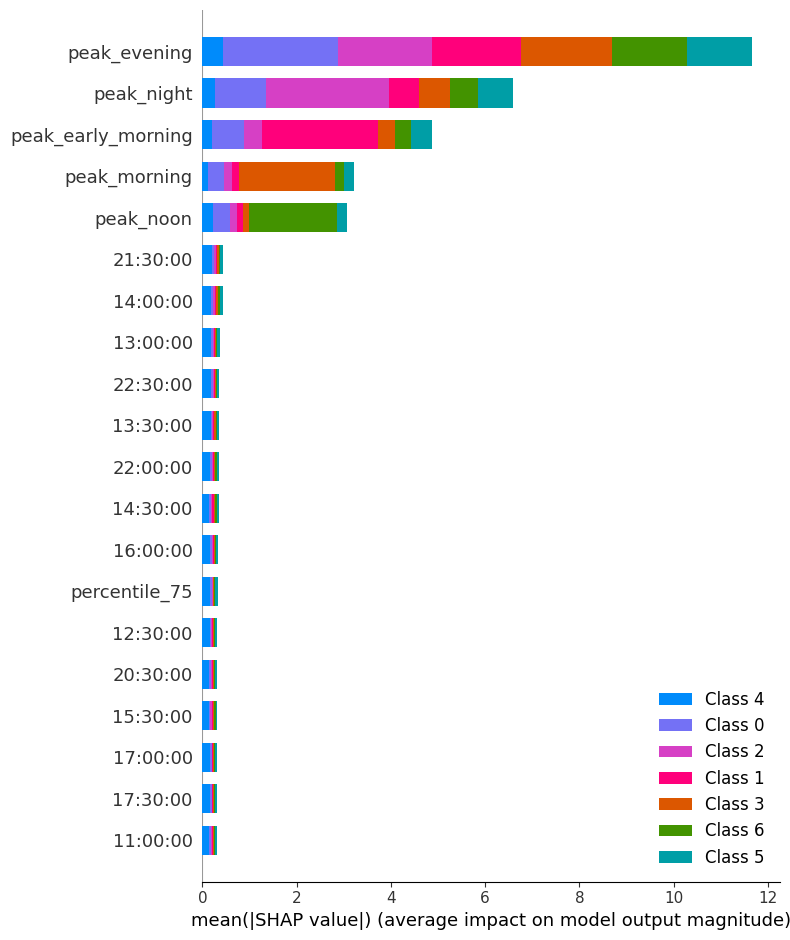}}
\caption{Average impact of each feature for all classes}
\label{fig:shap}
\end{figure*}

\begin{figure*}[ht!]
  \centering
  \includegraphics[width=0.6\linewidth]{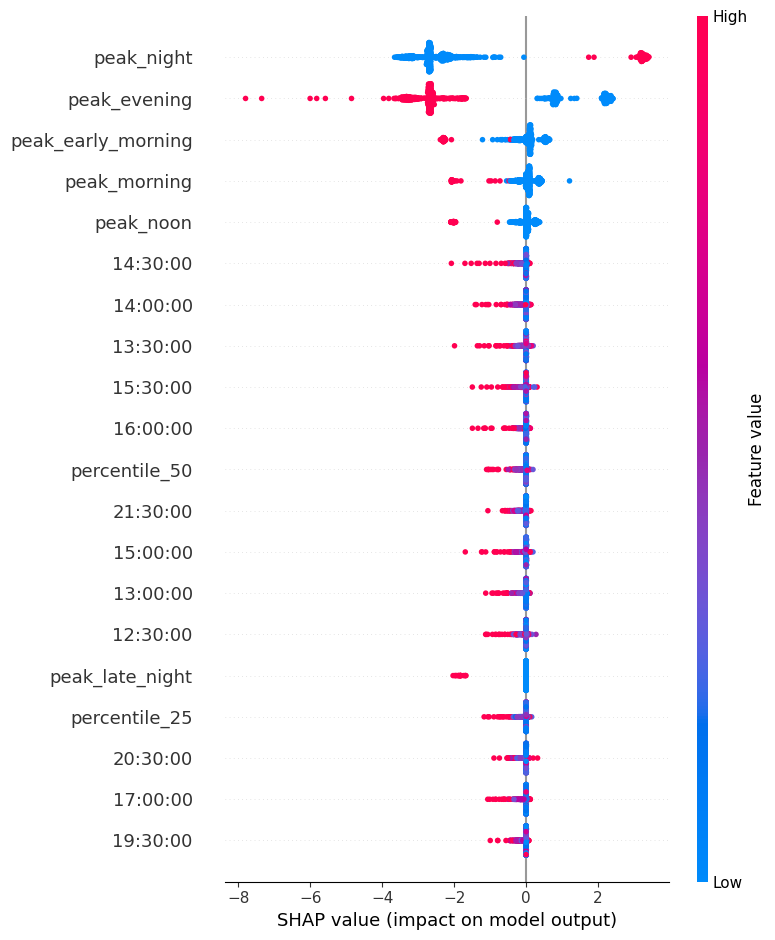}
  \caption{Average impact of each feature for a random household}
  \label{fig:shap_2}
\end{figure*}

Another way of validating our results is through xAI. Figure \ref{fig:shap}, depicts the average impact of each feature for all the classes for the whole testing dataset as well as Figure \ref{fig:shap_2} for a random household. This plot offers profound insights, thereby underscoring the substantial practical implications of our methodology in a business context. The association between each class and its respective peak hour is notably clear, shedding light on the classifier's rationale for grouping them together within a single class.

Furthermore, for programs and initiatives related to data-driven decision-making or targeted interventions, the insights derived from this analysis can serve as a crucial foundation. Specifically, in the case of DR programs, the information gathered from the feature importance analysis can guide the development of more effective and tailored interventions to consumers. By understanding which features have the most significant influence on a user's profile being classified into a specific group, program designers can focus their efforts on addressing those key factors.

\begin{figure*}[ht!]
    \centering
    \begin{subfigure}[b]{0.49\textwidth}
        \centering
        \includegraphics[width=\textwidth]{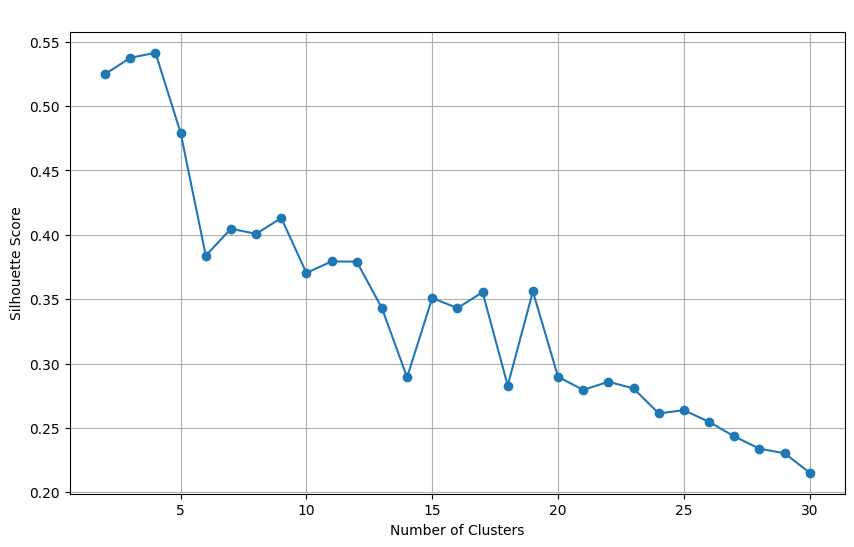}
        \caption{K-means}
        \label{fig:subfigure13}
    \end{subfigure}
    \begin{subfigure}[b]{0.49\textwidth}
        \centering
        \includegraphics[width=\textwidth]{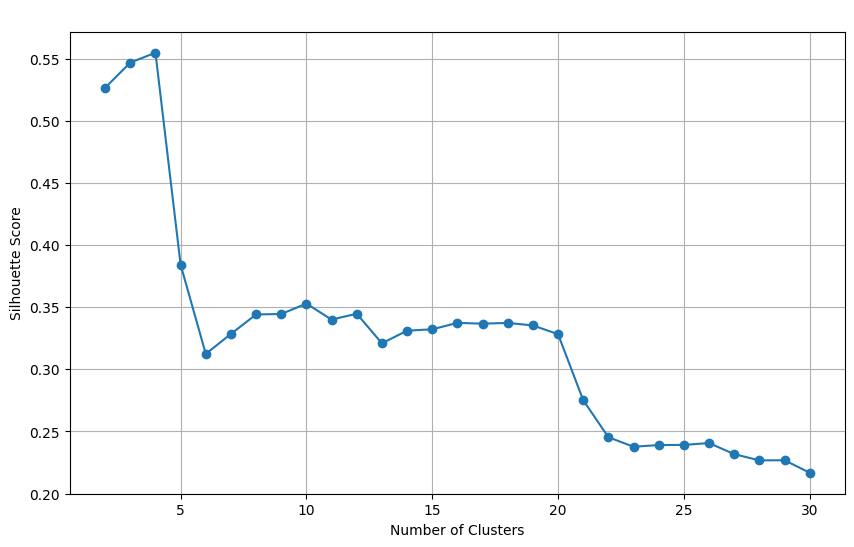}
        \caption{Agglomerative hierarchical}
        \label{fig:subfigure33}
    \end{subfigure}
    \caption{Silhouette score for every number of clusters and algorithm}
    \label{fig:shilouette_combined_2}
\end{figure*}

In this step having the aforementioned knowledge about the clusters produced, we can restart our pipeline for the user profiles that were assigned to clusters 4 and 5. For simplicity we are going to reach only the second level of our methodology, constructing the clusters in the first place. As we can see in Figure \ref{fig:shilouette_combined_2} and in Table \ref{tab:experiment_results_2} those two classes can be split in half when processed alone. Obviously, K-medoids is not a part of this analysis as it where excluded in earlier steps. Following this, the density based algorithm again produces 4 classes for the best indexes when tested for different $e$ values.

\begin{table}[ht!]
\footnotesize
\centering
\caption{Optimal number of clusters for each algorithm based on three evaluation indexes}
\begin{tabular}{|c|c|c|c|}
\hline
\textbf{Clustering Algorithm} & \textbf{Silhouette} & \textbf{DBI} & \textbf{CHI} \\
\hline
K-means & 4 & 4 & 3 \\
\hline
Agglomerative Clustering & 4 & 4 & 4 \\
\hline
\end{tabular}
\label{tab:experiment_results_2}
\end{table}

As previously mentioned, these user profiles do not align with any of the pre-established clusters. Consequently, this reevaluation gives birth to the emergence of two new user groups. These newly constructed clusters collectively represent a significant portion, approximately $10\%$, of the entire dataset. This finding not only highlights the flexibility and depth of our clustering approach but also suggests that our methodology can adapt to the evolving nature of the data, accommodating new patterns and subgroups as they emerge. This adaptability is a valuable asset in electrical user profiling, as it ensures that our pipeline remains relevant and effective in capturing the inherent complexity of user behavior over time.

\section{Conclusions}
\label{sec:conclusions}

In this paper, we address a critical challenge in the realm of electrical load segmentation and by extension DR programs, which play a pivotal role in achieving sustainable and resilient energy systems. The efficient management of electricity consumption in households is of paramount importance in the contemporary energy landscape, and the segmentation of electricity consumers into distinct groups based on their consumption patterns and behaviors is a fundamental requirement for the successful implementation of DR programs. Given the lack of mature information systems to support such assessments, several projects are currently straggling to receive financial support by funding aggregators to incorporate smart meters in order to maximize the knowledge of residential electrical load consumption. 

In this respect, a data-driven methodology is proposed which aspires to pave the way for aggregators and electricity providers to formulate and adjust DR programs based on their business and environmental goals. The proposed framework includes unsupervised learning techniques, namely clustering algorithms, that based on household electricity trends analysis, can select the optimal number of user groups based on some related indexes. Furthermore, the results are then evaluated through similarity matrices between the algorithms, selecting the nominal groups and the ones that need more attention. Finally, the problem is reincarnated as a probabilistic classification one making way for xAI techniques to give further business insights into the reason why these clusters where created in the first place. 

To expound upon this further, it is crucial to delve into the scalability implications of our approach, particularly in light of integrating the classifier component. Scalability concerns can encompass several facets, including the capacity to handle larger datasets, the computational efficiency required for processing, and the adaptability to a broader range of applications and use cases. The methodology is evaluated considering a real case study in London.In conclusion, the clustering algorithm analysis indicates that the ideal number of clusters in this scenario is seven. However, our methodology reveals that approximately 10\% of the dataset comprises two clusters with notable internal dissimilarities. As a result, we further subdivide them, resulting in a total of nine clusters.

Incorporating these insights into DR programs can lead to enhanced user outcomes, cost-effectiveness, and overall program efficiency. It allows for a more precise allocation of resources and interventions to individuals based on the specific factors that influence their classification, ultimately improving the program's ability to achieve its objectives. While acknowledging the limitations of this study, it is imperative that future research in the field of electricity user segmentation places a heightened emphasis on incorporating a comprehensive range of factors. These factors should encompass not only behavioral and socio-demographic aspects but also delve into the intricate realm of psychological factors that influence consumer choices and behaviors. By integrating these dimensions into our segmentation approach, we can construct more completed and meaning full user profiles for DR programs. More importantly, the proposed approach ought to become more relevant for large-scale applications and datasets. The inclusion of real-time data, coupled with robust forecasting capabilities for predicting electricity consumption trends in the coming days, would significantly reduce the complexity and limitations of the current methodology.

\section*{Acknowledgments}

The work presented is based on research conducted within the framework of the Horizon Europe European Commission project DEDALUS (Grant Agreement No. 101103998). The content of the paper is the sole responsibility of its authors and does not necessary reflect the views of the EC.

\bibliographystyle{unsrt}
\bibliography{annot}

\begin{thebibliography}{10}

\bibitem{4275494}
M.~H. Albadi and E.~F. El-Saadany.
\newblock Demand response in electricity markets: An overview.
\newblock In {\em 2007 IEEE Power Engineering Society General Meeting}, pages
  1--5, 2007.

\bibitem{Bahrami2012}
Sh. Bahrami, M.~Parniani, and A.~Vafaeimehr.
\newblock A modified approach for residential load scheduling using smart
  meters.
\newblock In {\em IEEE PES Innovative Smart Grid Technologies Conference
  Europe}, 2012.

\bibitem{Parvania20131957}
Masood Parvania, Mahmud Fotuhi-Firuzabad, and Mohammad Shahidehpour.
\newblock Optimal demand response aggregation in wholesale electricity markets.
\newblock {\em IEEE Transactions on Smart Grid}, 4(4):1957 – 1965, 2013.
\newblock Cited by: 271.

\bibitem{Muratori20161108}
Matteo Muratori and Giorgio Rizzoni.
\newblock Residential demand response: Dynamic energy management and
  time-varying electricity pricing.
\newblock {\em IEEE Transactions on Power Systems}, 31(2):1108 – 1117, 2016.
\newblock Cited by: 294.

\bibitem{en15051659}
Daiva Stanelyte, Neringa Radziukyniene, and Virginijus Radziukynas.
\newblock Overview of demand-response services: A review.
\newblock {\em Energies}, 15(5), 2022.

\bibitem{KAUR2022109236}
Ramanpreet Kaur and Dušan Gabrijelčič.
\newblock Behavior segmentation of electricity consumption patterns: A cluster
  analytical approach.
\newblock {\em Knowledge-Based Systems}, 251:109236, 2022.

\bibitem{Qiu2023}
Dawei Qiu, Yi~Wang, Junkai Wang, Chuanwen Jiang, and Goran Strbac.
\newblock Personalized retail pricing design for smart metering consumers in
  electricity market.
\newblock {\em Applied Energy}, 348, 2023.
\newblock Cited by: 0; All Open Access, Hybrid Gold Open Access.

\bibitem{Shi2023142}
Mengge Shi, Han Wang, Peng Xie, Cheng Lyu, Linni Jian, and Youwei Jia.
\newblock Distributed energy scheduling for integrated energy system clusters
  with peer-to-peer energy transaction.
\newblock {\em IEEE Transactions on Smart Grid}, 14(1):142 – 156, 2023.
\newblock Cited by: 3.

\bibitem{Qiu2021}
Dawei Qiu, Yujian Ye, Dimitrios Papadaskalopoulos, and Goran Strbac.
\newblock Scalable coordinated management of peer-to-peer energy trading: A
  multi-cluster deep reinforcement learning approach.
\newblock {\em Applied Energy}, 292, 2021.
\newblock Cited by: 53.

\bibitem{6693793}
Jungsuk Kwac, June Flora, and Ram Rajagopal.
\newblock Household energy consumption segmentation using hourly data.
\newblock {\em IEEE Transactions on Smart Grid}, 5(1):420--430, 2014.

\bibitem{Wang20193125}
Yi~Wang, Qixin Chen, Tao Hong, and Chongqing Kang.
\newblock Review of smart meter data analytics: Applications, methodologies,
  and challenges.
\newblock {\em IEEE Transactions on Smart Grid}, 10(3):3125 – 3148, 2019.
\newblock Cited by: 673; All Open Access, Green Open Access.

\bibitem{MOTLAGH201911}
Omid Motlagh, Adam Berry, and Lachlan O'Neil.
\newblock Clustering of residential electricity customers using load time
  series.
\newblock {\em Applied Energy}, 237:11--24, 2019.

\bibitem{en11092235}
Zigui Jiang, Rongheng Lin, and Fangchun Yang.
\newblock A hybrid machine learning model for electricity consumer
  categorization using smart meter data.
\newblock {\em Energies}, 11(9), 2018.

\bibitem{RAJABI2020109628}
Amin Rajabi, Mohsen Eskandari, Mojtaba~Jabbari Ghadi, Li~Li, Jiangfeng Zhang,
  and Pierluigi Siano.
\newblock A comparative study of clustering techniques for electrical load
  pattern segmentation.
\newblock {\em Renewable and Sustainable Energy Reviews}, 120:109628, 2020.

\bibitem{1626400}
G.~Chicco, R.~Napoli, and F.~Piglione.
\newblock Comparisons among clustering techniques for electricity customer
  classification.
\newblock {\em IEEE Transactions on Power Systems}, 21(2):933--940, 2006.

\bibitem{CHICCO201268}
Gianfranco Chicco.
\newblock Overview and performance assessment of the clustering methods for
  electrical load pattern grouping.
\newblock {\em Energy}, 42(1):68--80, 2012.
\newblock 8th World Energy System Conference, WESC 2010.

\bibitem{MCLOUGHLIN2015190}
Fintan McLoughlin, Aidan Duffy, and Michael Conlon.
\newblock A clustering approach to domestic electricity load profile
  characterisation using smart metering data.
\newblock {\em Applied Energy}, 141:190--199, 2015.

\bibitem{1295037}
G.~Chicco, R.~Napoli, F.~Piglione, P.~Postolache, M.~Scutariu, and C.~Toader.
\newblock Load pattern-based classification of electricity customers.
\newblock {\em IEEE Transactions on Power Systems}, 19(2):1232--1239, 2004.

\bibitem{LOPEZ2011716}
José~J. López, José~A. Aguado, F.~Martín, F.~Muñoz, A.~Rodríguez, and
  José~E. Ruiz.
\newblock Hopfield–k-means clustering algorithm: A proposal for the
  segmentation of electricity customers.
\newblock {\em Electric Power Systems Research}, 81(2):716--724, 2011.

\bibitem{1304160}
G.~Chicco, R.~Napoli, and F.~Piglione.
\newblock Application of clustering algorithms and self organising maps to
  classify electricity customers.
\newblock In {\em 2003 IEEE Bologna Power Tech Conference Proceedings,},
  volume~1, pages 7 pp. Vol.1--, 2003.

\bibitem{YILMAZ2019665}
S.~Yilmaz, J.~Chambers, and M.K. Patel.
\newblock Comparison of clustering approaches for domestic electricity load
  profile characterisation - implications for demand side management.
\newblock {\em Energy}, 180:665--677, 2019.

\bibitem{5484234}
S.~M. Bidoki, N.~Mahmoudi-Kohan, M.~H. Sadreddini, M.~Zolghadri~Jahromi, and
  M.~P. Moghaddam.
\newblock Evaluating different clustering techniques for electricity customer
  classification.
\newblock In {\em IEEE PES T\&D 2010}, pages 1--5, 2010.

\bibitem{article}
João Soares, Fernando Lezama, Tiago Pinto, and Hugo Morais.
\newblock Complex optimization and simulation in power systems.
\newblock {\em Complexity}, 2018:1--3, 10 2018.

\bibitem{8056062}
Amin Rajabi, Li~Li, Jiangfeng Zhang, Jianguo Zhu, Sahand Ghavidel, and
  Mojtaba~Jabbari Ghadi.
\newblock A review on clustering of residential electricity customers and its
  applications.
\newblock In {\em 2017 20th International Conference on Electrical Machines and
  Systems (ICEMS)}, pages 1--6, 2017.

\bibitem{4524366}
Florentin Batrinu, Gianfranco Chicco, Roberto Napoli, Federico Piglione, Petru
  Postolache, Mircea Scutariu, and Cornel Toader.
\newblock Efficient iterative refinement clustering for electricity customer
  classification.
\newblock In {\em 2005 IEEE Russia Power Tech}, pages 1--7, 2005.

\bibitem{9072418}
Nameer Al~Khafaf, Mahdi Jalili, and Peter Sokolowski.
\newblock A novel clustering index to find optimal clusters size with
  application to segmentation of energy consumers.
\newblock {\em IEEE Transactions on Industrial Informatics}, 17(1):346--355,
  2021.

\bibitem{6607329}
Ioannis~P. Panapakidis, Minas~C. Alexiadis, and Grigoris~K. Papagiannis.
\newblock Deriving the optimal number of clusters in the electricity consumer
  segmentation procedure.
\newblock In {\em 2013 10th International Conference on the European Energy
  Market (EEM)}, pages 1--8, 2013.

\bibitem{7338120}
Hanna Schäfer, Joaquim~L. Viegas, Marta~C. Ferreira, Susana~M. Vieira, and
  J.~M.~C. Sousa.
\newblock Analysing the segmentation of energy consumers using mixed fuzzy
  clustering.
\newblock In {\em 2015 IEEE International Conference on Fuzzy Systems
  (FUZZ-IEEE)}, pages 1--7, 2015.

\bibitem{8833693}
Ahmad~Khaled Zarabie, Sahar Lashkarbolooki, Sanjoy Das, Kumarsinh Jhala, and
  Anil Pahwa.
\newblock Load profile based electricity consumer clustering using affinity
  propagation.
\newblock In {\em 2019 IEEE International Conference on Electro Information
  Technology (EIT)}, pages 474--478, 2019.

\bibitem{7279510}
Yanlong Wang, Li~Li, and Qinmin Yang.
\newblock Application of clustering technique to electricity customer
  classification for load forecasting.
\newblock In {\em 2015 IEEE International Conference on Information and
  Automation}, pages 1425--1430, 2015.

\bibitem{6688044}
Jungsuk Kwac, Chin-Woo Tan, Nicole Sintov, June Flora, and Ram Rajagopal.
\newblock Utility customer segmentation based on smart meter data: Empirical
  study.
\newblock In {\em 2013 IEEE International Conference on Smart Grid
  Communications (SmartGridComm)}, pages 720--725, 2013.

\bibitem{https://doi.org/10.1002/asmb.2453}
Emilie Devijver, Yannig Goude, and Jean-Michel Poggi.
\newblock Clustering electricity consumers using high-dimensional regression
  mixture models.
\newblock {\em Applied Stochastic Models in Business and Industry},
  36(1):159--177, 2020.

\bibitem{kmeans}
David Arthur and Sergei Vassilvitskii.
\newblock K-means++ the advantages of careful seeding.
\newblock In {\em Proceedings of the eighteenth annual ACM-SIAM symposium on
  Discrete algorithms}, pages 1027--1035, 2007.

\bibitem{kmedoids}
Leonard Kaufman and Peter~J Rousseeuw.
\newblock {\em Finding groups in data: an introduction to cluster analysis}.
\newblock John Wiley \& Sons, 2009.

\bibitem{DBSCAN}
Martin Ester, Hans-Peter Kriegel, Jorg Sander, Xiaowei Xu, et~al.
\newblock A density-based algorithm for discovering clusters in large spatial
  databases with noise.
\newblock In {\em kdd}, volume~96, pages 226--231, 1996.

\bibitem{DBI}
David Davies and Don Bouldin.
\newblock A cluster separation measure.
\newblock {\em Pattern Analysis and Machine Intelligence, IEEE Transactions
  on}, PAMI-1:224 -- 227, 05 1979.

\bibitem{CBI}
T.~Caliński and J~Harabasz.
\newblock A dendrite method for cluster analysis.
\newblock {\em Communications in Statistics}, 3(1):1--27, 1974.

\bibitem{sne}
Laurens Van~der Maaten and Geoffrey Hinton.
\newblock Visualizing data using t-sne.
\newblock {\em Journal of machine learning research}, 9(11), 2008.

\bibitem{catboost}
Liudmila Prokhorenkova, Gleb Gusev, Aleksandr Vorobev, Anna~Veronika Dorogush,
  and Andrey Gulin.
\newblock Catboost: unbiased boosting with categorical features.
\newblock {\em Advances in neural information processing systems}, 31, 2018.

\end{thebibliography}

\clearpage


\end{document}